%% file: neurips_2020.tex
\documentclass{article}

     \PassOptionsToPackage{numbers, compress}{natbib}

     \usepackage[final]{neurips_2020}

\usepackage[utf8]{inputenc} 
\usepackage[T1]{fontenc}    
\usepackage{hyperref}       
\usepackage{url}            
\usepackage{booktabs}       
\usepackage{amsmath}
\usepackage{amsfonts}       
\usepackage{nicefrac}       
\usepackage{microtype}      

\usepackage[pdftex]{graphicx} 
\usepackage{adjustbox}

\usepackage{tikz}
\usepackage{pgfplots}
\pgfplotsset{compat=1.16}
\usepackage{pgfplotstable}
\usepackage{bm}

\usepackage{multirow}
\usepackage{textcomp}  

\usetikzlibrary{arrows, decorations.pathmorphing, decorations.pathreplacing, positioning, calc}

\definecolor{olivegreen}{rgb}{0,0.6,0}

\usepackage{color}

\usepackage[final]{changes}
\title{Improving Natural Language Processing Tasks with Human Gaze-Guided Neural Attention}

\author{Ekta Sood$^1$, Simon Tannert$^2$, Philipp M\"uller$^1$, Andreas Bulling$^1$\\
$^1$University of Stuttgart, Institute for Visualization and Interactive Systems (VIS), Germany \\
$^2$University of Stuttgart, Institute for Natural Language Processing (IMS), Germany \\
\texttt{\{ekta.sood, philipp.mueller, andreas.bulling\}@vis.uni-stuttgart.de}\\
\texttt{simon.tannert@ims.uni-stuttgart.de}
}

\begin{document}

\maketitle

\begin{abstract}
A lack of corpora has so far limited advances in integrating human gaze data as a supervisory signal in neural attention mechanisms for natural language processing (NLP).
We propose a novel \textit{hybrid text saliency model} (TSM) that, for the first time, combines a cognitive model of reading with explicit human gaze supervision in a single machine learning framework.
On four different corpora we demonstrate that our hybrid TSM duration predictions are highly correlated with human gaze ground truth.
We further propose a novel \textit{joint modeling approach} to integrate TSM predictions into the attention layer of a network designed for a specific upstream NLP task without the need for any task-specific human gaze data.
We demonstrate that our joint model outperforms the state of the art in paraphrase generation on the Quora Question Pairs corpus by more than 10\% in BLEU-4 and achieves state of the art performance for sentence compression on the challenging Google Sentence Compression corpus.
As such, our work introduces a practical approach for bridging between data-driven and cognitive models and demonstrates a new way to integrate human gaze-guided neural attention into NLP tasks.
\end{abstract}

\section{Introduction}

Neural attention mechanisms have been widely applied in  computer vision and have been shown to enable neural networks to only focus on those aspects of their input that are important for a given task~\cite{mnih2014recurrent,xu2015show}.
While neural networks are able to learn meaningful attention mechanisms using only supervision received for the target task, the addition of human gaze information has been shown to be beneficial in many cases~\cite{karessli2017gaze,qiao2018exploring,xu2015gaze,yun2013studying}.
An especially interesting way of leveraging gaze information was demonstrated by works incorporating human gaze into neural attention mechanisms, for example for image and video captioning~\cite{sugano2016seeing,yu2017supervising} or visual question answering~\cite{qiao2018exploring}.

While attention is at least as important for reading text as it is for viewing images~\cite{commodari2005attention,wolfe2017five}, integration of human gaze into neural attention mechanisms for natural language processing (NLP) tasks remains under-explored.
A major obstacle to studying such integration is data scarcity:
Existing corpora of human gaze during reading consist of too few samples to provide effective supervision for modern data-intensive architectures and human gaze data is only available for a small number of NLP tasks.
For paraphrase generation and sentence compression, which play an important role for tasks such as reading comprehension systems~\cite{gupta2018deep,hermann2015teaching,patro2018learning}, no human gaze data is available.

We address this data scarcity in two novel ways:
First, to overcome the low number of human gaze samples for reading, we propose a novel hybrid text saliency model (TSM) in which we combine a cognitive model of reading behavior with human gaze supervision in a single machine learning framework.
More specifically, we use the E-Z Reader model of attention allocation during reading~\cite{reichle1998toward} to obtain a large number of synthetic training examples.
We use these examples to pre-train a BiLSTM~\cite{graves2005framewise} network with a Transformer~\cite{vaswani2017attention} whose weights we subsequently refine by training on only a small amount of human gaze data.
We demonstrate that our model yields predictions that are well-correlated with human gaze on out-of-domain data.
Second, we propose a novel joint modeling approach of attention and comprehension that allows human gaze predictions to be flexibly adapted to different NLP tasks by integrating TSM predictions into an attention layer.
By jointly training the TSM with a task-specific network, the saliency predictions are adapted to this upstream task without the need for explicit supervision using real gaze data.
Using this approach, we outperform the state of the art in paraphrase generation on the Quora Question Pairs corpus by more than 10\% in BLEU-4 and achieve state of the art performance on the Google Sentence Compression corpus.
As such, our work demonstrates the significant potential of combining cognitive and data-driven models and establishes a general principle for flexible gaze integration into NLP that has the potential to also benefit tasks beyond paraphrase generation and sentence compression.

\section{Related work}

Our work is related to previous works on 1) NLP tasks for text comprehension, 2) human attention modeling, as well as 3) gaze integration in neural network architectures.

\subsection{NLP tasks for text comprehension}

Two key tasks in machine text comprehension are paraphrasing and summarization~\cite{chen2016thorough,hermann2015teaching,cho2019paraphrase,li2017paraphrase,gupta2010survey}.
While paraphrasing is the task of ``conveying the same meaning, but with different expressions''~\cite{cho2019paraphrase,fader2013paraphrase,li2017paraphrase}, summarization deals with extracting or abstracting the key points of a larger input sequence~\cite{frintrop2010computational,tas2007survey,kaushik2018much}.
Though advances have helped bring machine comprehension closer to human performance, humans are still superior for most tasks~\cite{blohm2018comparing,xia2019automatic,zhang2018record}.
While attention mechanisms can improve performance by helping models to focus on relevant parts of the input~\cite{prakash2016neural,rush2015neural,rocktaschel2015reasoning,cao2016attsum,hasan2016neural,cho2015describing}, the benefit of explicit supervision through human attention remains under-explored.

\subsection{Human attention modeling}

Predicting what people visually attend to in images (saliency prediction) is a long-standing challenge in neuroscience and computer vision~\cite{borji2012state,bylinskii2016should,kummerer2014deep}.
In contrast to images, most attention models for eye movement behaviors during reading are cognitive process models, i.e. models that do not involve machine learning but implement cognitive theories~\cite{engbert2005swift,rayner1978eye,reichle1998toward}.
Key challenges for such models are a limited number of parameters, hand-crafted rules and thus a difficulty to adapt them to different tasks and domains, as well as the difficulty to use them as part of an end-to-end trained machine learning architectures~\cite{duch2008cognitive,kotseruba201840,ma2020neural}.
One of the most influential cognitive models of gaze during reading is the E-Z Reader model~\cite{reichle1998toward}
It assumes attention shifts to be strictly serial in nature and that saccade production depends on different stages of lexical processing.
that has been successful in explaining different effects seen in attention allocation during reading~\cite{reichle2009using,reichle2013using}.

In contrast, learning-based attention models for text remain under-explored.
\citet{nilsson2009learning} trained person-specific models on features including length and frequency of words to predict fixations and later extended their approach to also predict fixation durations~\cite{nilsson2010towards}.
The first work to present a person-independent model for fixation prediction on text used a linear CRF model~\cite{matthies2013blinkers}.
A separate line of work has instead tried to incorporate assumptions about the human reading process into the model design.
For example, the Neural Attention Trade-off (NEAT) language model  was trained with hard attention and assigned a cost to each fixation \citet{hahn2016modeling}.
Subsequent work applied the NEAT model to question answering tasks, showing task-specific effects on learned attention patterns that reflect human behavior~\cite{DBLP:journals/corr/abs-1808-00054}.
Further approaches include sentence representation learning using surprisal and part of speech tags as proxies to human attention~\cite{wang2016learning}, attention as a way to improve time complexity for NLP tasks~\cite{seo2017neural}, and learning saliency scores by training for sentence comparison~\cite{samardzhiev2018learning}.
Our work is fundamentally different from all of these works in that we, for the first time, combine cognitive theory and data-driven approaches.

\subsection{Gaze integration in neural network architectures}

Integration of human gaze data into neural network architectures has been explored for a range of computer vision tasks~\cite{karessli2017gaze,shcherbatyi2015gazedpm,xu2015gaze,yu2017supervising,yun2013studying}.
\citet{sugano2016seeing} were the first to use gaze as an additional input to the attention layer for image captioning, while \citet{qiao2018exploring} used human-like attention maps as an additional supervision for the attention layer for a visual question answering task.
Most previous work in gaze-supported NLP has used gaze as an input feature, e.g. for syntactic sequence labeling~\cite{klerke2019glance}, classifying referential versus non-referential use of pronouns~\cite{yaneva2018classifying}, reference resolution~\cite{iida2011multi}, key phrase extraction~\cite{zhang2019using}, or prediction of multi-word expressions~\cite{rohanian2017using}. Recently, \citet{hollenstein2019advancing} proposed to build a lexicon of gaze features given word types, overcoming the need for gaze data at test time.
Two recent works proposed methods inspired by multi-task learning to integrate gaze into NLP classification tasks.~\cite{klerke2016improving} did not integrate gaze into the attention layers but demonstrated performance improvements by adding a gaze prediction task to regularize a sentence compression model.~\cite{barrett2018sequence} did not predict human gaze for the target task but used ground-truth gaze from another eye tracking corpus to regularize their neural attention function.
In stark contrast, our work is the first to combine a cognitive model of reading and a data-driven approach to predict human gaze, to directly integrate these predictions into the neural attention layers, and to jointly train for two different tasks -- generative (paraphrase generation) and classification (sentence compression).

\section{Method}

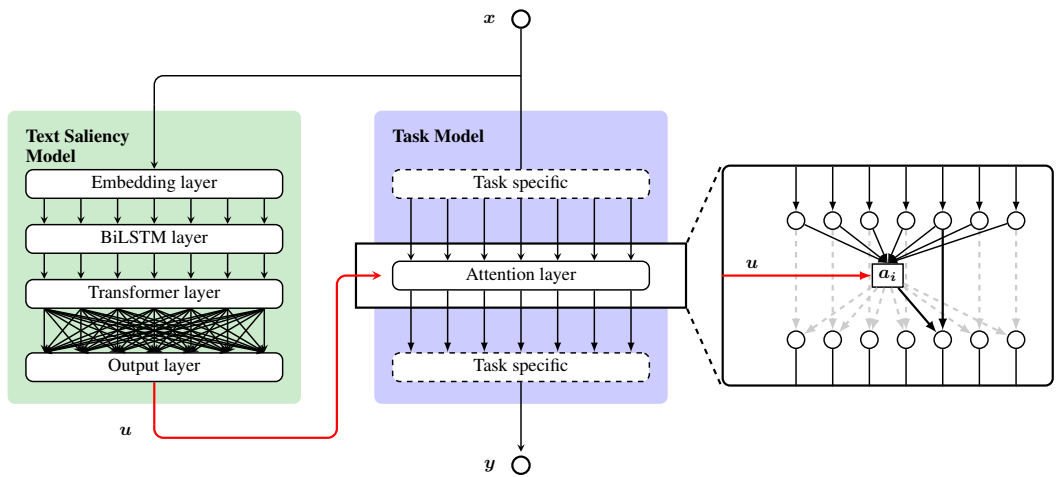
\begin{figure}[!t]
\centering
\adjustbox{width=\linewidth}{\input{figures/1}}
\caption{High-level architecture of our model. Given an input sentence $x_1...x_n$ the Text Saliency Model produces attention scores $u_1...u_n$ for each word in the input sentence $x$. The Task Model combines this information with the original input sentence to produce an output sentence $y_1...y_m$.}
\label{fig:NeuRIPS2020/figures/1}
\end{figure}

We make two contributions:
A hybrid text saliency model as well as two attention-based models for paraphrase generation and text summarization employed in a novel joint modeling approach\footnote{Code and other supporting material can be found at \url{https://perceptualui.org/publications/sood20\_neurips/}}.

\subsection{Hybrid text saliency model}

To overcome the limited amount of eye-tracking data for reading comprehension tasks, we propose \textit{a hybrid approach} when training our text saliency model.
In the first stage of training, we leverage the E-Z Reader model~\cite{reichle1998toward} to generate a large amount of training data over the CNN and Daily Mail Reading Comprehension Corpus~\cite{hermann2015teaching}.
After training the text saliency model until convergence using this synthetic data, in a second training phase we fine-tune the network with real eye tracking data of humans reading from the Provo and Geco corpus~\cite{luke2018provo,cop2017presenting}.
We used the most recent implementation of EZ Reader (Version 10.2) available from the authors' website\footnote{\url{http://www.erikdreichle.com/downloads.html}}.

The task of text saliency is to predict fixation durations $u_i$ for each word $x_i$ of an input sentence.
In our text saliency model, we combine a BiLSTM network~\cite{graves2005framewise} with a Transformer~\cite{vaswani2017attention} (cf.~\autoref{fig:NeuRIPS2020/figures/1} for an overview).
Each word $x_i$ of the input sentence is encoded using pre-trained GloVe embeddings~\cite{pennington2014glove}.
The resulting embeddings are fed into a single-layer BiLSTM network~\cite{graves2005framewise} that integrates information over the whole input sentence.
The outputs from the BiLSTM network are fed into a Transformer network with multi-headed self-attention~\cite{vaswani2017attention}.
In contrast to~\cite{vaswani2017attention}, we only use the encoder of the Transformer network.
Furthermore, we do not provide positional encodings as input because this information is already implicitly present in the outputs of the BiLSTM layer.
In preliminary experiments we found advantages in using only four layers with four attention heads each for the Transformer network in contrast to the six layers with 12 heads in the original architecture~\cite{vaswani2017attention}.
We also found the combination of a BiLSTM network with a subsequent Transformer network to yield predictions that are most similar to human data, particularly with longer sequence lengths.
According to~\cite{wang2019r}, this might be due to the transformer encoding coarse relational information about positions of sequence elements, while the BiLSTM better captures fine-grained word level context.
The specific choice of architecture therefore allows our model to better capture the sequential context while still maintaining computational efficiency.
Finally, a fully connected layer is used to obtain an attention score $u_i$ for each input word $x_i$ in $x$.
We apply sigmoid nonlinearities with subsequent normalization over the input sentence to obtain a probability distribution over the sentence.
As loss function we use the mean squared error.

\subsection{Joint modeling for natural language processing tasks}

To model the relationship between attention allocation and text comprehension, we integrate the TSM with two different NLP task attention-based networks in a \textit{joint model} (cf.~\autoref{fig:NeuRIPS2020/figures/1}).
Specifically, we propose a modification to the Luong attention layer~\cite{DBLP:journals/corr/LuongPM15} that is a computationally light-weight but highly effective, multiplicative attention algorithm~\cite{DBLP:journals/corr/LuongPM15,britz2017massive}.
We compute attention scores $a_i$ as

\begin{equation}
    a_i = \text{softmax}(\text{score}_\text{T}(h_i, s_j))
\end{equation}

using our task-specific modified score functions $\text{score}_\text{T}$.
For the tasks of paraphrase generation and sentence compression, respectively, we propose the novel score functions

\begin{equation}
    \label{eq:score_paragen}
    \text{score$_\text{ParaGen}$}(h_i, s_j) = u \odot h_i^\top W_as_j
\end{equation}

\begin{equation}
    \text{score$_\text{TextComp}$}(h_i, s_j) = u \odot v_a^\top\tanh(W_a[h_i;s_j])
    \label{eq:score_textComp}
\end{equation}

Where $h_i$ is the current hidden state, $s_j$ are the hidden states of the encoder and $W_a$ and $v_a$ are learnable parameters of the attention mechanism.
The outputs of the TSM model $u$ on the input sentence are incorporated into the score function by element-wise multiplication.
This way, attention scores in the upstream task network reflect word saliencies learnt from humans. In addition to that, the error signal from the upstream loss function can be propagated back to the TSM in order to adapt its' parameters to the upstream task, thereby defining an implicit loss on $u$.
This way, the attention distribution $u$ returned by the TSM is adapted to the specific upstream task, allowing us to incorporate and adapt a neural model of attention to tasks for which no human gaze data is available.
Note, as we have two different tasks namely generative (paraphrase generation) and classification (sentence compression), we used different score functions as suggested by previous work~\cite{DBLP:journals/corr/LuongPM15}.

\section{Experiments} 

\subsection{Joint model with upstream tasks}

\subsubsection*{Evaluation details}

\paragraph{Datasets} We used two standard benchmark corpora to evaluate each upstream NLP task.
For paraphrase generation, we used the Quora Question Pairs corpus\footnote{\url{https://www.quora.com/q/quoradata/First-Quora-Dataset-Release-Question-Pairs}} that consists of human-annotated pairs of paraphrased questions that were crawled from Quora.
We followed the common practice of excluding negative paraphrase examples from the corpus to obtain training data for paraphrase generation~\cite{patro2018learning,gupta2018deep}.
We split the data according to~\cite{gupta2018deep,patro2018learning}, using either 100K or 50K examples for training, 45K examples for validation, and 4K examples for testing. 
For the sentence compression task we used the Google Sentence Compression corpus~\cite{filippova2015sentence} containing 200K sentence compression pairs that were crawled from news articles.
We split the data according to~\cite{zhao2018language}, taking the first 1K examples as test data, and the next 1K as validation data.

\paragraph{Paraphrase generation} Our first text comprehension task was paraphrase generation where, given a source sentence, the model has to produce a different target sentence with the same meaning that may have a different length.
We used a sequence-to-sequence network with word-level attention that was originally proposed for neural machine translation~\cite{bahdanau2014neural}.
The model consisted of two recurrent neural networks, an encoder and an attention decoder (cf.~\autoref{fig:NeuRIPS2020/figures/1}).
The encoder consisted of an embedding layer followed by a gated recurrent unit (GRU)~\cite{cho2014learning}.
The decoder produced an output sentence step-by-step given the hidden state of the encoder and the input sentence.
At each output step, the encoded input word and the previous hidden state are used to produce attention weights using our modified Luong attention (cf.~\autoref{eq:score_paragen}).
These attention weights are combined with the embedded input sentence and fed into a GRU to produce an output sentence.
The loss between predicted and the ground-truth paraphrase was calculated over the entire vocabulary using cross-entropy.

\paragraph{Sentence compression} As a second task, we opted for deletion-based sentence compression that aims to delete unimportant words from an input sentence~\cite{jing2000sentence,knight2002summarization,mcdonald2006discriminative,clarke2008global,filippova2015sentence}.
We incorporated the attention mechanism into the baseline architecture presented in~\cite{filippova2015sentence}.
The network consisted of three stacked LSTM layers with dropout after each LSTM layer as a regularization method.
The outputs of the last LSTM layer were fed through our modified Luong attention mechanism (cf.~\autoref{eq:score_textComp}) and two fully connected layers which predicted for each word whether it should be deleted.
The loss between predicted and ground truth deletion mask was calculated with cross-entropy.

\paragraph{Training} We used pre-trained 300-dimensional GloVe embeddings in both the TSM and the upstream task network to represent the input words~\cite{pennington2014glove}.
We trained both upstream task models using the ADAM optimizer~\cite{kingma2014adam} with a learning rate of 0.0001. 
For paraphrase generation we used uni-directional GRUs with hidden layer size 1,024 and dropout probability of 0.2. For sentence compression we used BiLSTMs with hidden layer size 1,024 and dropout probability of 0.1.

\paragraph{Metrics} The most common metric to evaluate text generative tasks is BLEU~\cite{papineni2002bleu}, which measures the n-gram overlap between the produced and target sequence.
To ensure reproducibility, we followed the standard Sacrebleu~\cite{post2018call} implementation that uses BLEU-4.
For sentence compression, we followed previous works~\cite{filippova2015sentence,zhao2018language} by reporting the F1 score as well as the compression ratio calculated as the length of the compressed sentence divided by the input sentence length measured in characters~\cite{filippova2015sentence}.

\subsubsection*{Results and discussion}

\begin{table}[!t]
    \caption{Ablation study results and comparison with the state of the art for paraphrase generation with both data splits in terms of BLEU-4 score for different training set sizes and sentence compression in terms of F1 score and compression ratio.
    Also shown is the number of model parameters.}
    \label{tab:JMPG_Metrics}
    \centering
    \adjustbox{width=\linewidth}{
    \begin{tabular}{llllllll}
        \toprule
        \multicolumn{4}{c}{Paraphrase Generation (BLEU-4)} & 
        \multicolumn{4}{c}{Sentence Compression} \\
        \cmidrule(r){1-4}
        \cmidrule(r){5-8}
        Method & 50K & 100K & Params & Method & F1 & CR & Params \\
        \midrule
        \added{& & & & Klerke et al. (2016) & 80.9 & --- & ---} \\
        Baseline (Seq-to-Seq) & 7.11 & 8.91 & 45M & Baseline (BiLSTM) & 81.3 & 0.39 & 12M \\
        Patro et al. (2018) & 16.5 & 17.9 & --- & Zhao et al. (2018) & \textbf{85.1} & 0.39 & --- \\
        \midrule
        No Fixation & 24.62 & 27.81 & 69M & No Fixation & 83.4 & 0.38 & 129M \\
        Random TSM Init & 25.26 & 27.11 & 79M & Random TSM Init & 83.7 & 0.38 & 178M \\
        TSM Weight Swap & 23.43 & 27.60 & 79M & TSM Weight Swap & 83.8 & 0.38 & 178M \\
        Frozen TSM & 25.73 & 28.26 & 79M & Frozen TSM & 83.9 & 0.37 & 178M\\
        Ours & \textbf{26.24} & \textbf{28.82} & 79M & Ours & \textbf{85.0} & 0.39 & 178M \\
         \bottomrule
    \end{tabular}}
\end{table}

Results for our joint model on paraphrase generation and sentence compression in comparison to the state of the art are shown in ~\autoref{tab:JMPG_Metrics}.
As can be seen in the table, for paraphrase generation our approach achieves a BLEU-4 score of 28.82 when using 100K training examples, clearly outperforming the previous state of the art for this task from~\cite{patro2018learning} (17.9 BLEU-4).
The same holds for 50K training examples (26.24 vs.\ 16.5 BLEU-4).
For sentence compression, our joint model achieves a F1 score of 85.0 and a compression rate of 0.39.
This is on par with the state of the art performance of 85.1 F1 score and 0.39 compression rate reported in~\citet{zhao2018language}\footnote{For a more detailed comparison to our model see table 1 in the supplementary material.}.
In that work, a syntax-based language model was used to learn the syntactic dependencies between lexical items in the given input sequence. 
In contrast, our current method does not require any syntax-based language model, but it will be interesting to see
whether it will benefit from additional syntactic information in future work.
When comparing our results for sentence compression on the Google dataset to~\cite{klerke2016improving} we observe an increase of \textasciitilde 5\% F1 score for our method (cf. Table~\ref{tab:JMPG_Metrics}).

To further analyze the impact of our joint modeling approach, we evaluated several ablated versions of our model:
\begin{itemize}
    \item \textbf{Baseline (Seq-to-Seq):} Stand-alone models based on a Seq-2-Seq network~\cite{bahdanau2014neural} for paraphrase generation and a BiLSTM network~\cite{schuster1997bidirectional} for sentence compression.
    \item \textbf{No Fixations:} Stand-alone upstream task network with original Luong attention (no TSM).
    \item \textbf{Random TSM Init:} Random initialization of the TSM instead of training on E-Z Reader and human data. Still implicit supervision by the upstream task during joint training.
    \item \textbf{TSM Weight Swap:} Exchange of the weights of the TSM model between tasks, i.e. sentence compression using the TSM weights obtained from the best-performing paraphrase generation model and vice versa.
    \item \textbf{Frozen TSM:} Training of the TSM with E-Z Reader and human gaze predictions but with frozen weights in the joint training with the upstream task, i.e. no adaptation of the TSM.
\end{itemize}

\begin{figure}[!t]
\centering
\includegraphics[width=1\linewidth]{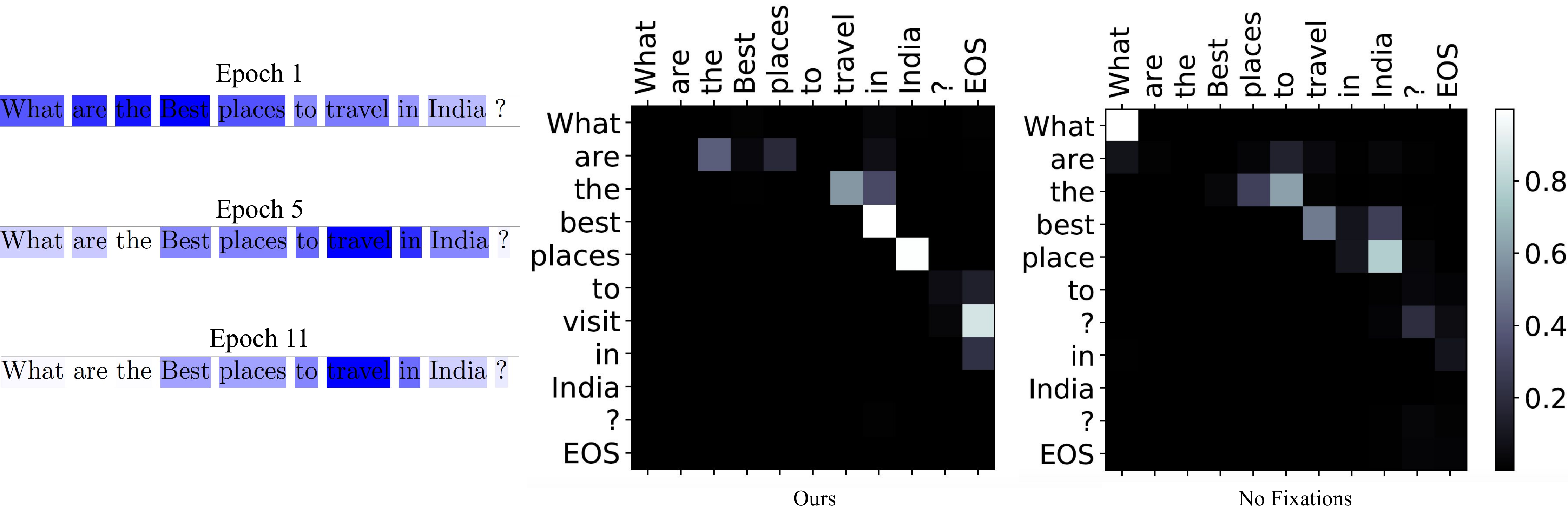}
\caption{Paraphrase generation attention maps for both sub-networks (TSM predictions and upstream task attention) in our joint architecture. We show the TSM fixation predictions (left in blue) over epochs (last epoch is our converged models). We show the two-dimensional neural attention maps (right), showing our model and the \textit{no fixation} model from our ablation study. The two-dimensional maps show the input sequence (horizontal axis) and the predicted sequence (vertical axis). We show the temporal TSM predictions over epochs, in order to depict how the fixation predictions change while training. The fixation predictions (for each epoch, left) are computed over words in the input sequences and then are integrated into the neural attention mechanism which in turn is used to make a prediction (vertical axis, right).}
\label{fig:Adapted_Saliency_To_TaskPG}
\end{figure}

\begin{figure}[!t]
\centering
\includegraphics[width=1\linewidth]{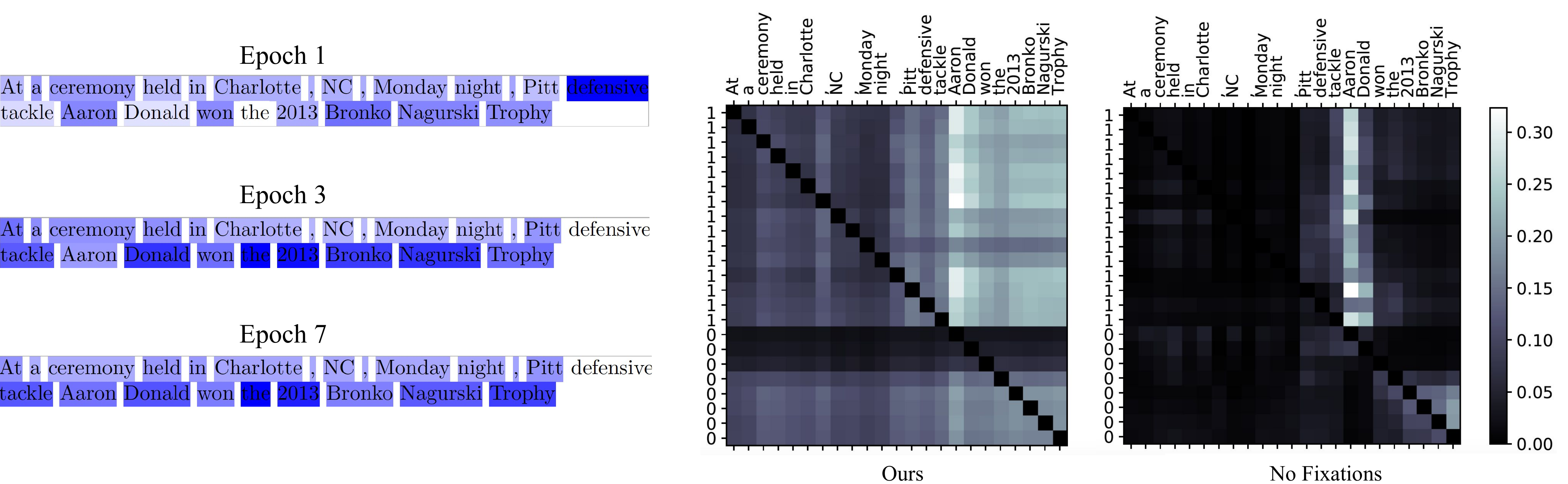}
\caption{Sentence compression attention maps for both sub-networks (TSM predictions and upstream task attention) in our joint architecture. We show the TSM fixation predictions (left in blue) over epochs (last epoch is our converged models). We show the two-dimensional neural attention maps (right), showing our model and the \textit{no fixation} model from our ablation study. The two-dimensional maps show the input sequence (horizontal axis) and the predicted sequence (vertical axis). We show the temporal TSM predictions over epochs, in order to depict how the fixation predictions change while training. The fixation predictions (for each epoch, left) are computed over words in the input sequences and then are integrated into the neural attention mechanism which in turn is used to make a prediction (vertical axis, right).}
\label{fig:Adapted_Saliency_To_TaskSC}
\end{figure}

As can be seen from~\autoref{tab:JMPG_Metrics}, all ablated models obtain inferior performance to our full model on both tasks (statistically significant at the 0.05 level).
Notably, even the \textit{No Fixation} model improves drastically over the \textit{Seq-to-Seq baseline} for paraphrase generation, most likely due to the significant increase in network parameters.
The benefit of training the \textit{TSM} with our hybrid approach before using it in the joint model is underlined by the performance difference between the \textit{Random TSM Init} (e.g.\ decrease in performance for both tasks) and our full model (e.g.\ best performance and differently adapted saliency predictions (cf.~\autoref{tab:JMPG_Metrics} and~\autoref{fig:Adapted_Saliency_To_TaskPG},~\autoref{fig:Adapted_Saliency_To_TaskSC}).\footnote{Additional 1D and 2D maps over all conditions are available in the supplementary material.}

Most importantly, our full model achieves higher performance than the \textit{Frozen TSM} model in all evaluations (e.g.\ 85.0 vs.\ 83.9 F1 for sentence compression), indicating that our model successfully adapts the TSM predictions during joint training.
This is further underlined by the inferior performance of the \textit{TSM Weight Swap} model:
Swapping the optimal TSM weights between different upstream tasks leads to a notable performance decrease (e.g.\ 85.0 vs.\ 83.7 F1 for sentence compression), implying that the TSM model adaptation is specific to the upstream task.

To gain insights into how our joint model training adapts TSM predictions to specific upstream tasks, we analyzed the saliency predictions over time. \autoref{fig:Adapted_Saliency_To_TaskPG} and~\autoref{fig:Adapted_Saliency_To_TaskSC} show visualizations of representative samples for both tasks over multiple training epochs.
As can be seen in the left half of the figures, the adapted saliency predictions differ significantly from each other.
In paraphrase generation (cf.~\autoref{fig:Adapted_Saliency_To_TaskPG}), the saliency predictions focus on fewer words in the sentence within 11 epochs, specifically the word ``travel'' that is replaced in the correct paraphrase by ``visit''.
For sentence compression (cf.~\autoref{fig:Adapted_Saliency_To_TaskSC}), the predictions continue to be spread over the whole sentence with only slight changes in the distribution over the words.
This makes sense given that the task of this network is to delete as many words in the input sequence as possible while still maintaining syntactic structure and meaning.

The right half of the figures show 2D neural attention maps of the converged models with the input sequence on the horizontal and the prediction on the vertical axis for \textit{our} (with fixations) and the \textit{No Fixation} model, respectively.
As can be seen, our model correctly predicts the paraphrase, while the \textit{No Fixation} model does not.
Also, both the converged models neural attention weights differ with respect to allocation of probability mass.
We see the \textit{No Fixation} model densely concentrates attention towards a specific few input words (horizontal axis) when predicting several words (vertical axis).
In contrast, the attention mass of our model is more spread out.

\subsection{Pre-training of the hybrid text saliency model (TSM)}

\begin{figure}[!t]
\centering
\includegraphics[width=\linewidth]{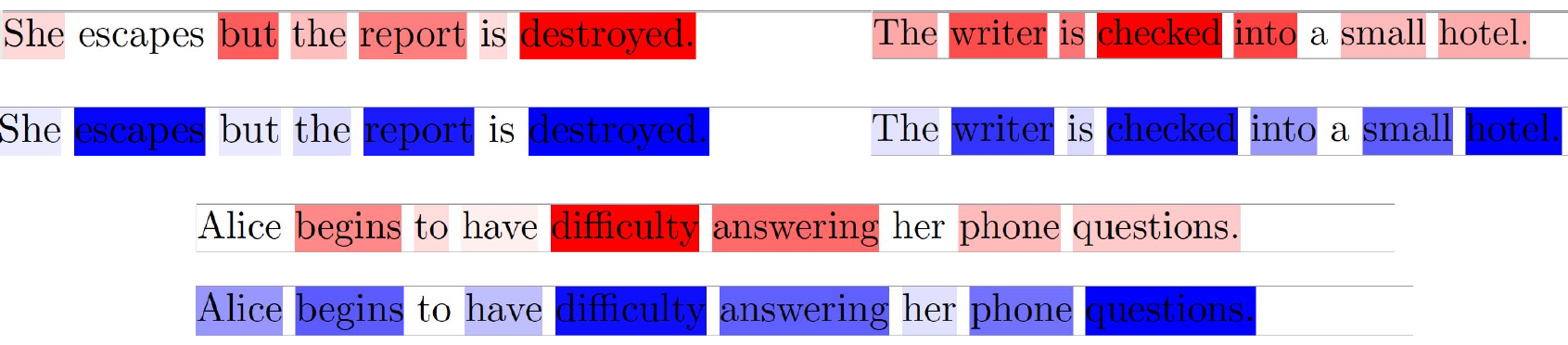}
\caption{Heatmaps showing human fixation durations in red and hybrid TSM duration predictions in blue. Here we show three different example sentences in order to depict the similarity between TSM word-level durations predictions as compared to human ground truth word-level durations} 
\label{fig:saliency_model_heatmaps}
\end{figure} 

\subsubsection*{Evaluation details}

\paragraph{Training datasets} Training the TSM consists of two stages: pre-training with synthetic data generated by E-Z Reader, and subsequent fine-tuning on human gaze data.
For the first step, we run E-Z Reader on the CNN and Daily Mail corpus~\cite{hermann2015teaching} consisting of 300K online news articles with on average 3.75 sentences.
As recommended in~\citet{reichle1998toward}, we run E-Z Reader 10 times for each sentence to ensure stability in fixation predictions.
For training we obtain a total of 7.6M annotated sentences on Daily Mail and 3.1M for CNN. For validation, we obtained 850K sentences on Daily Mail and 350K on CNN.
For the second step, we used the two established gaze corpora Provo~\cite{luke2018provo} and Geco~\cite{cop2017presenting}.
Provo contains 55 short passages, extracted from different sources such as popular science magazines and fiction stories~\cite{luke2018provo}.
We split the data into 10K sentence pairs (pairs means sentence to human, as multiple humans read the same sentence) for train and 1K sentence pairs for validation.
Geco is comprised of long passages from a popular novel~\cite{cop2017presenting}. 
We split the data into 65K sentence pairs for train and 8K sentence pairs for validation.

\paragraph{Test datasets} We evaluated our model on the validation sets of the Provo and Geco corpora, as well as on the Dundee~\cite{kennedy2005parafoveal} and MQA-RC corpora~\cite{sood20_conll}.
The combined validation corpora of Provo and Geco contained 18K sentence pairs. 
Dundee consists of recordings from 10 participants reading 20 news articles while MQA-RC corpus is a 3-condition reading comprehension corpus using 32 documents from the MovieQA question answering dataset~\cite{MovieQA}.
For our evaluation we used 1K sentence pairs from the free reading condition.
This dataset is substantially different from the other eye tracking corpora because its stimuli are scraped from online sources and contain noise not found in text intended for human reading.

\paragraph{Implementation details} We used pre-trained 300 dimensional GloVe word embeddings~\cite{pennington2014glove}.
Our network has a bidirectional LSTM, with four transformer self-attention layers with four heads and hidden size of 128.
The model objective is to predict normalized fixation durations for each word in the input sentence, resulting in saliency scores between 0 and 1.
We used the ADAM optimizer~\cite{kingma2014adam} with a learning rate of 0.00001, batch size of 100, and dropout of 0.5 after the embedding layer and the recurrent layer. 
We pre-trained our network on synthetic training data for four epochs, and then fine-tune it on human data for 10 epochs.

\begin{table}
    \caption{Comparison of predicted and human ground-truth fixation durations for the different TSM conditions and corpora in terms of mean squared error (MSE), Jensen Shannon Divergence (JSD), and Spearman's rank correlation ($\rho$) between the part of speech tags based fixation distributions for model predictions and ground truth. 
    A star indicates statistically significant $\rho$ at $p < 0.05$.}
  \label{tab:REGRESSION_Metrics}
    \centering
    \begin{tabular}{llllllllll}
        \toprule
        & \multicolumn{3}{c}{TSM} & \multicolumn{3}{c}{TSM w/o pre-training} & \multicolumn{3}{c}{TSM w/o fine-tuning} \\
        \cmidrule(r){2-4}
        \cmidrule(r){5-7}
        \cmidrule(r){8-10}
        Corpus & MSE & JSD & $\rho$ & MSE & JSD & $\rho$ & MSE & JSD & $\rho$ \\
        \midrule
        Dundee & 0.063 & 0.39 & 0.99* & 0.071 & 0.39 & 0.99* & 0.096 & 0.47 & -0.68 \\
        Provo + Geco & 0.105 & 0.34 & 1.00* & 0.112 & 0.36 & 0.99* & 0.238 & 0.46 & 0.10 \\
        Provo & 0.003 & 0.24 & 0.88* & 0.008 & 0.44 & 0.83* & 0.032 & 0.52 & -0.25 \\
        Geco & 0.118 & 0.35 & 0.99* & 0.127 & 0.35 & 0.98* & 0.267 & 0.45 & -0.10 \\
        MQA-RC & 0.064 & 0.36 & 0.94* & 0.071 & 0.36 & 0.76* & 0.083 & 0.42& -0.05\\
        \bottomrule
    \end{tabular}
\end{table}

\paragraph{Metrics} To evaluate the TSM model, we compute mean squared error (MSE) between the predicted and ground truth fixation durations as well as the Jensen-Shannon Divergence (JSD)~\cite{lin1991divergence}.
JSD is widely used in eye tracking research to evaluate inter-gaze agreement~\cite{mozaffari2018evaluating,fang2009between,davies2016exploring,oertel2013gaze} as, unlike Kullback-Leibler Divergence, JSD is symmetric.
In addition we measured the word type predictability as it is a well-known predictor of fixation probabilities~\cite{hahn2016modeling,nilsson2009learning}.
We used the Stanford tagger~\cite{10.3115/1073445.1073478} to predict part-of-speech (POS) tags for our corpora and compute the average fixation probability per tag,
allowing us to compute the correlation between our model and ground truth using Spearman's $\rho$.

\subsubsection*{Results and discussion}

~\autoref{tab:REGRESSION_Metrics} shows the performance of our model and ablation conditions in terms of means squared error (MSE), Jensen-Shannon-Divergence (JSD) and correlation to human ground truth.
As ablation conditions we evaluate a model only trained on human data (w/o pre-train) as well as a model that is not fine-tuned on human data (w/o fine-tune), but only trained with E-Z Reader data.

Most importantly, our model is superior to- or on par with both ablation variants across all metrics and corpora, showing the importance of both the E-Z Reader pre-training as well as the fine-tuning with human data.
Pre-training with data obtained from E-Z Reader is most beneficial in the case of the small Provo corpus, where we observe a reduction from 0.44 JSD to 0.24 JSD by adding the pre-training step.
For the larger corpora this difference is less pronounced but still present.
It is interesting to note that TSM w/o fine-tune performs consistently the worst, indicating that training on E-Z Reader data alone insufficient even though it provides benefits when combined with human data.

Using the correlations to human gaze over the POS distributions, we can compare our approach to~\citet{hahn2016modeling} who achieved a $\rho$ of 0.85 on the Dundee corpus, compared to a $\rho$ of 0.99 achieved by our model.
Furthermore we observe an especially large improvement in $\rho$ as a result of E-Z Reader pre-training on the MQA-RC dataset.
This dataset, unlike the other eye tracking corpora, is generated from stimuli which were scraped from online sources regarding movie plots, underlining the effectiveness of our approach in generalizing to out-of-domain data.
In further analyses on the POS based correlations we observed that content words, such as adjectives, adverbs, nouns, and verbs, are more predictive than function words.~\footnote{Detailed POS distributions are available in the supplementary material.}
Lastly, we provide a qualitative impression of our method by comparing attention maps using our TSM predictions to ground truth human data (cf. Figure~\ref{fig:saliency_model_heatmaps}).

\section{Conclusion}

In this work we made two novel contributions towards improving natural language processing tasks using human gaze predictions as a supervisory signal.
First, we introduced a novel hybrid text saliency model that, for the first time, integrates a cognitive reading model with a data-driven approach to address the scarcity of human gaze data on text.
Second, we proposed a novel joint modeling approach that allows the TSM to be flexibly adapted to different NLP tasks without the need for task-specific ground truth human gaze data.
We showed that both advances result in significant performance improvements over the state of the art in paraphrase generation as well as competitive performance for sentence compression but with a much less complex model than the state of the art.
We further demonstrated that this approach is effective in yielding task-specific attention predictions.
Taken together, our findings not only demonstrate the feasibility and significant potential of combining cognitive and data-driven models for NLP tasks -- and potentially beyond -- but also how saliency predictions can be effectively integrated into the attention layer of task-specific neural network architectures to improve performance.

\begin{ack}

E. Sood was funded by the Deutsche Forschungsgemeinschaft (DFG, German Research Foundation) under Germany's Excellence Strategy - EXC 2075 -- 390740016;
S. Tannert was supported by IBM Research AI through the IBM AI Horizons Network;
P. M\"uller and A. Bulling were funded by the European Research Council (ERC; grant agreement 801708). Additional revenues related to, but not supporting, this work: Scholarship by Google for E. Sood.
We would like to thank the following people for their helpful insights and contributions: Sean Papay, Pavel Denisov, Prajit Dhar, Manuel Mager, Diego Frassinelli, Fabian Koegel and Keerthana Jaganathan.

\end{ack}

\newpage

\bibliographystyle{plainnat}
\bibliography{neurips_2020}

\appendix

\section{Appendix}

\subsection{Sentence Compression Comparison To Previous SOTA}

To gain further insight into the comparison between our model and the current state of the art in sentence compression, we show results of our method and ablations in relation to ablations of the method by \citet{zhao2018language} (see Table~\ref{tab:Zhao_Comparision}).
In their work, the authors added a ``syntax-based language model'' to their sentence compression network with which they obtained the state-of-the-art performance of 85.1 F1 score. 
The authors employ a syntax-based language model which is trained to learn the syntactic dependencies between lexical items in the given input sequence. Together with this language model, they use a reinforcement learning algorithm to improve the deletion proposed by their Bi-LSTM model.
Using a naive language model without syntactic features their model obtained a F1 score of 85.0. 
With their stand-alone Bi-LSTM method in which they do not employ the reinforce language model policy, they obtain 84.8.
In contrast, our method does neither include a reinforcement-learning based language model nor additional syntactic features.
However, our method is still competitive with the state of the art (achieving a F1 score of 85.0), and
arguably might benefit from additional incorporation of syntactic information in future work. 

\begin{table}[!ht]
    \caption{Ablation study results and comparison with the state of the art for sentence compression generation in terms of F1 score and compression ratio.
    Also shown is the number of model parameters. We show that our model, \textit{without additional syntactic information} as was used in previous methods, still obtains SOTA performance.}
    \label{tab:Zhao_Comparision}
    \centering
    \begin{tabular}{lllll}
        \toprule
        & Method & F1 & CR & Params \\
        \midrule
        \multirow{3}{*}{Zhao et al (2018)} & LSTM implementation & 84.8 & 0.40 & --- \\
        & Evaluator LM & 85.0 & 0.41 & --- \\
        & Syntax-Based Evaluator LM & \textbf{85.1} & 0.39 & --- \\
        \midrule
        \multirow{6}{*}{Our paper} & Baseline (BiLSTM) & 81.3 & 0.39 & 12M \\
        & No Fixation & 83.4 & 0.38 & 129M \\
        & Random TSM Init & 83.7 & 0.38 & 178M \\
        & TSM Weight Swap & 83.8 & 0.38 & 178M \\
        & Frozen TSM & 83.9 & 0.37 & 178M\\
        & Ours & \textbf{85.0} & 0.39 & 178M \\
         \bottomrule
    \end{tabular}
\end{table}

\subsection{Ablation Study -- Attention Maps}
\begin{figure}[!ht]
\centering
\includegraphics[width=1\linewidth]{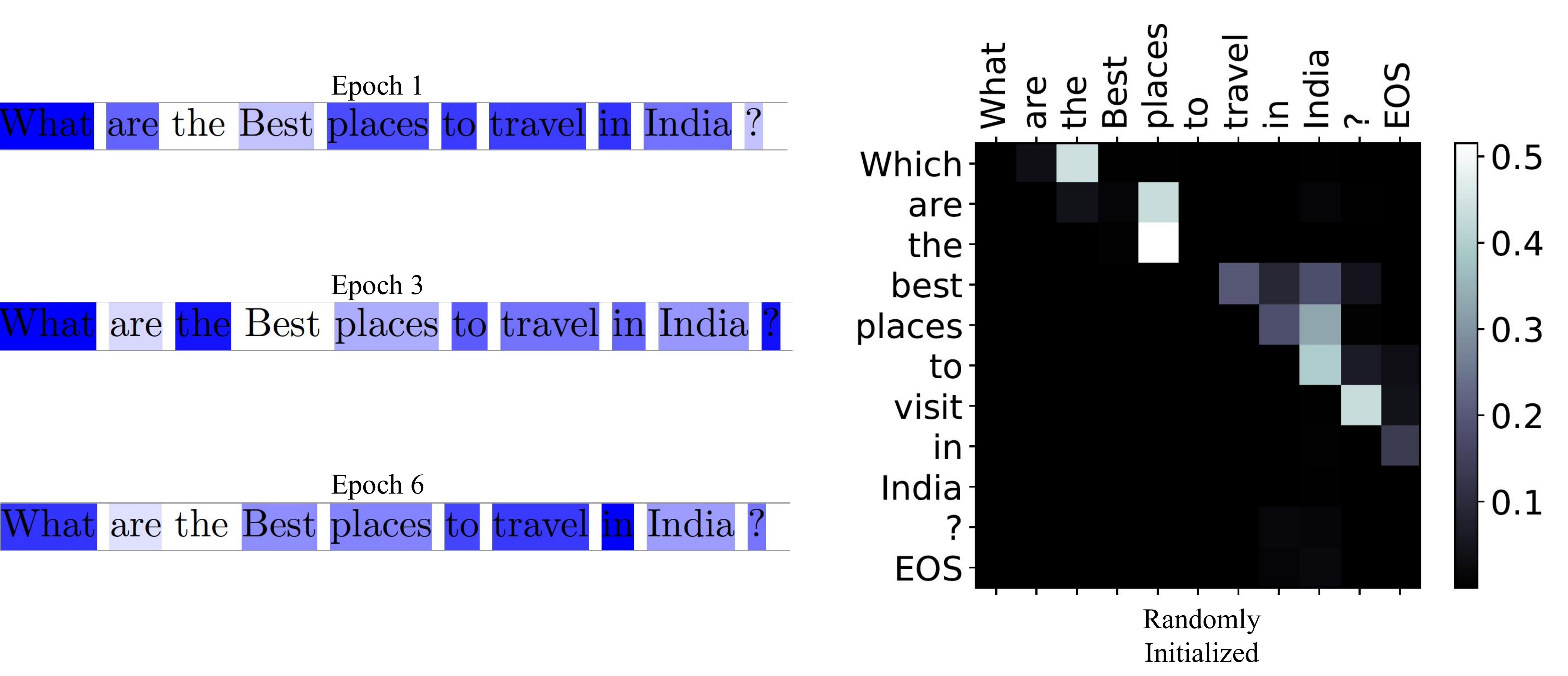}
\includegraphics[width=1\linewidth]{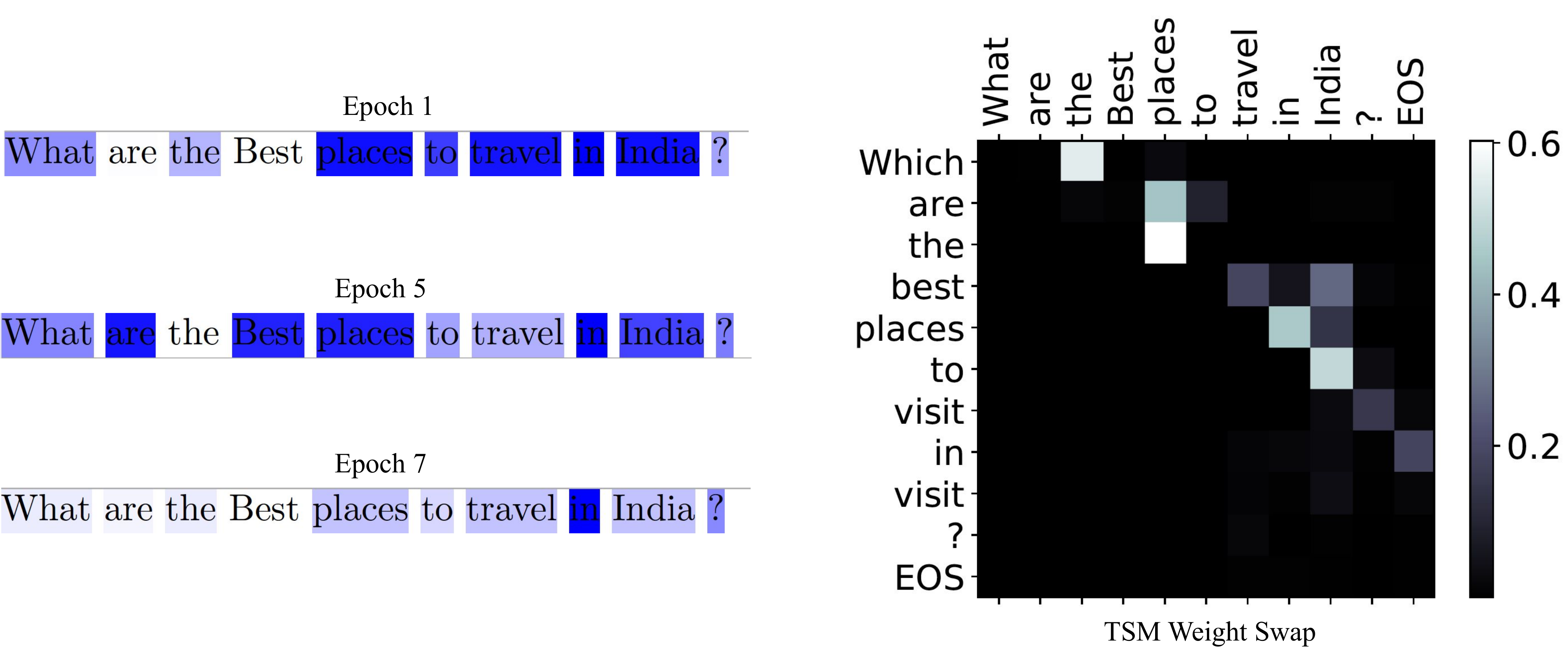}
\caption{Additional paraphrase generation attention maps from our ablation study, for both sub-networks (TSM predictions and upstream task attention) in our joint architecture. We show the TSM fixation predictions (left in blue) over epochs (last epoch is our converged models). We show the two-dimensional neural attention maps (right), showing the \textit{Random TSM Init} (top) and \textit{TSM Weight Swap} (bottom) model from our ablation study. The two-dimensional maps show the input sequence (horizontal axis) and the predicted sequence (vertical axis). 
We show the temporal TSM predictions over epochs, in order to depict how the fixation predictions change while training.
The fixation predictions (for each epoch, left) are computed over words in the input sequences and then are integrated into the neural attention mechanism which in turn is used to make a prediction (vertical axis, right).
}
\label{fig:additional_PG_maps}
\end{figure}

\begin{figure}[!ht]
\centering
\includegraphics[width=1\linewidth]{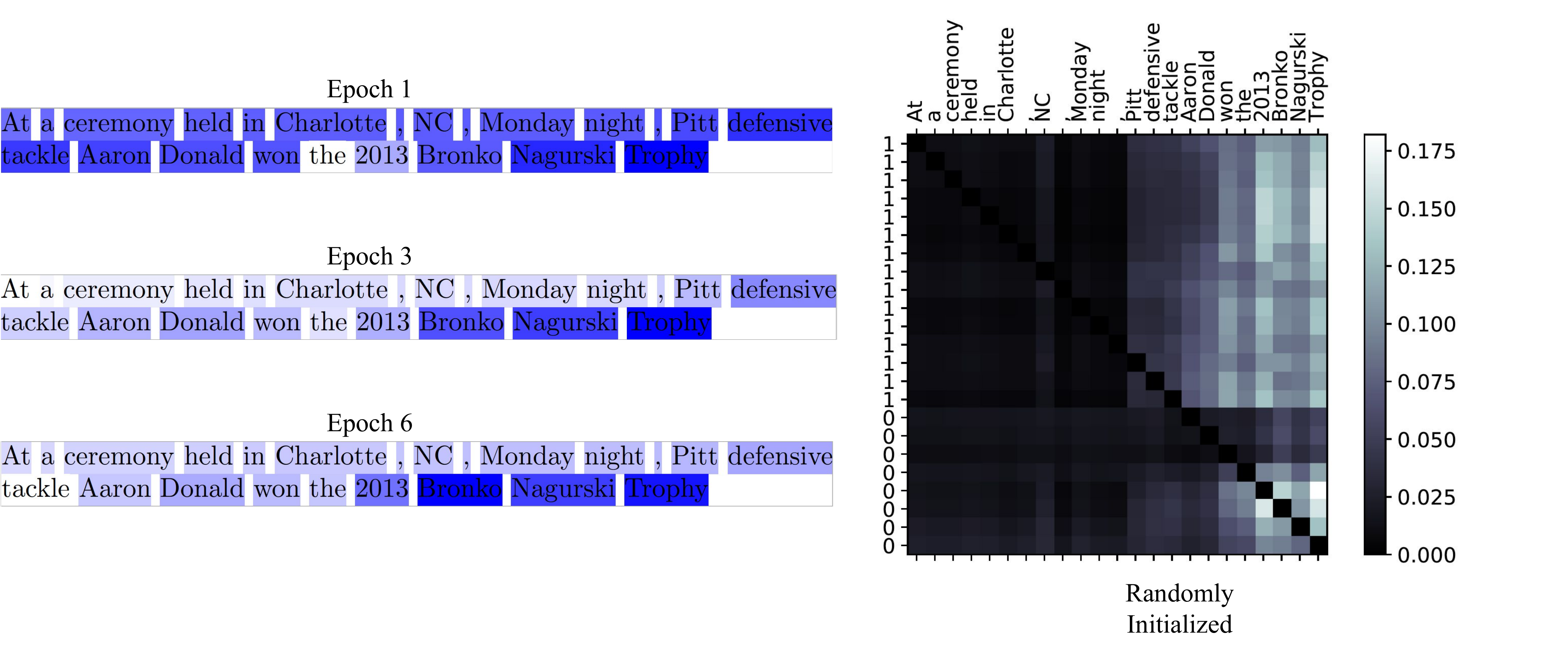}
\includegraphics[width=1\linewidth]{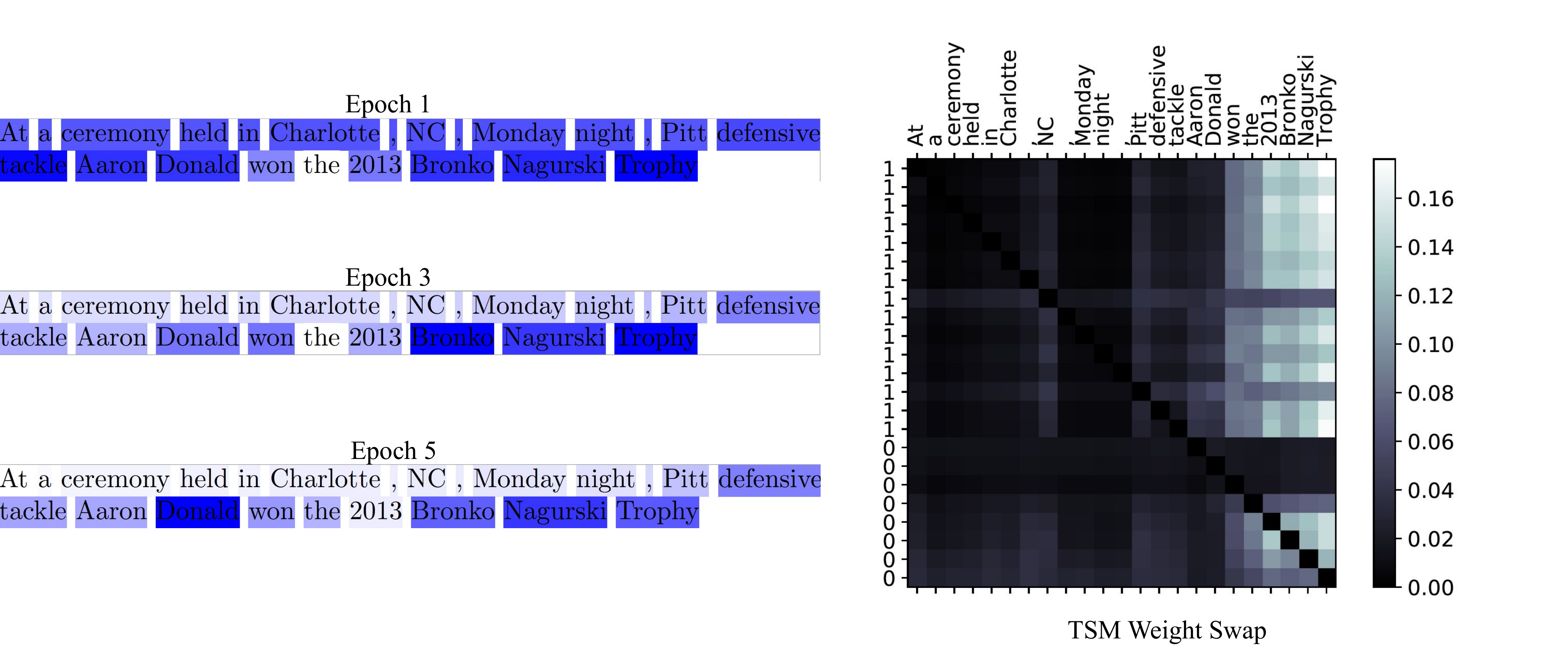}
\caption{Additional sentence compression attention maps from our ablation study, for both sub-networks (TSM predictions and upstream task attention) in our joint architecture. We show the TSM fixation predictions (left in blue) over epochs (last epoch is our converged models). We show the two-dimensional neural attention maps (right), showing the \textit{Random TSM Init} (top) and \textit{TSM Weight Swap} (bottom) model from our ablation study. The two-dimensional maps show the input sequence (horizontal axis) and the predicted sequence (vertical axis). We show the temporal TSM predictions over epochs, in order to depict how the fixation predictions change while training. The fixation predictions (for each epoch, left) are computed over words in the input sequences and then are integrated into the neural attention mechanism which in turn is used to make a prediction (vertical axis, right).}
\label{fig:additional_SC_maps}
\end{figure}

To shed more light onto the adapted TSM predictions for the conditions in our ablation study, we present saliency and neural attention maps for the conditions \textit{Random TSM Init} and \textit{TSM Weight Swap}. 
In Figure~\ref{fig:additional_PG_maps}, we show that the adapted saliency predictions (blue, left showing) for paraphrase generation, between the two conditions (top vs. bottom) vary with respect to the words which are predicted to be most salient and the temporal adaptation during training. The last epoch is from the converged models, respectively. 
There exist notable differences in the adapted TSM predictions for the two ablations.
However, we assume they do not play a role in performance between these two conditions, as these performance differences are not statistically significant. 
However, these conditions do perform significantly worse than our model (see paper for results).
As shown in the paper, our model allocates the most attention to the word ``travel'' in the example sentence.
This is the word that is changed in the paraphrase output, indicating that the our adapted TSM can effectively guide the paraphrase generation system.
Figure~\ref{fig:additional_SC_maps} shows the adapted saliency predictions for the sentence compression task. 
The differences between both conditions are less distinct, with minimal temporal variation in the word saliency predictions. As with the paraphrase generation models, performance differences between the two ablations are not statistically significant. 
Compared to the saliency output for our model (shown in the paper), we observe that our model more equally allocates attention to the part of the sentence that is going to be deleted.

While the 2d neural attention maps for the example sentence in the paraphrase generation task are similar for \textit{Random TSM Init} and \textit{TSM Weight Swap}, they differ clearly from the corresponding neural attention maps for our model (shown in the paper).
Similarly, the 2d neural attention maps for sentence compression (Figure~\ref{fig:additional_SC_maps}, right) are rather similar for \textit{Random TSM Init} and \textit{TSM Weight Swap}.
However, the corresponding neural attention map for our method presented in the paper is more spread out and additionally allocates more attention on the position in the input sentence from which on the network decides to delete words.
Taken together, these results illustrate the differences in neural attention that are connected to the superior performance of our full model over the ablation conditions.

\subsection{Part of Speech Distributions -- Content vs Function Words}

\begin{figure}[!ht]
\centering
\pgfplotstableread{
corpus function content
dundee(human) 0.36229852838121934 0.6377014716187807
dundee(model) 0.38608822103732426 0.6139117789626758
provo(human) 0.49584816132858833 0.5041518386714117
provo(model) 0.4473684210526316 0.5526315789473685
geco(human) 0.37932467532467534 0.6206753246753247
geco(model) 0.36527478106358674 0.6347252189364132
mqa(human) 0.4040601210228382 0.5959398789771617
mqa(model) 0.4130530973451328 0.5869469026548673
}\tabledata

\begin{tikzpicture} 

  \begin{axis}[
    width=.6\textwidth,
    height=240pt,
    ybar stacked,
    ymin=0,
    xtick=data,
    legend style={cells={anchor=west}, legend pos=outer north east},
    reverse legend=true,
    xlabel=Corpus (human vs. model),
    xlabel shift = 1.0cm,
    ylabel=Ratio (content vs. function words),
    xticklabels from table={\tabledata}{corpus},
    xticklabel style={rotate=-60,anchor=west,text width=1cm,align=center},
    set layers,  
    extra y ticks=.5, 
    extra y tick labels={},
    extra y tick style={
        ymajorgrids=true,
        ytick style={
            /pgfplots/major tick length=0pt,
        },
        grid style={
            black,
            dashed,
            /pgfplots/on layer=axis foreground,  
        },
    },
    ]
    \addplot table [y=content, meta=corpus, x expr=\coordindex] {\tabledata};
    \addlegendentry{content}
    \addplot table [y=function, meta=corpus, x expr=\coordindex] {\tabledata};
    \addlegendentry{function}
  \end{axis}
\end{tikzpicture}
    \caption{Per-sentence normalized gaze durations on content words versus function words for our TSM model and human gaze data across different corpora.
    }
    \label{fig:POS_distribution}
\end{figure}
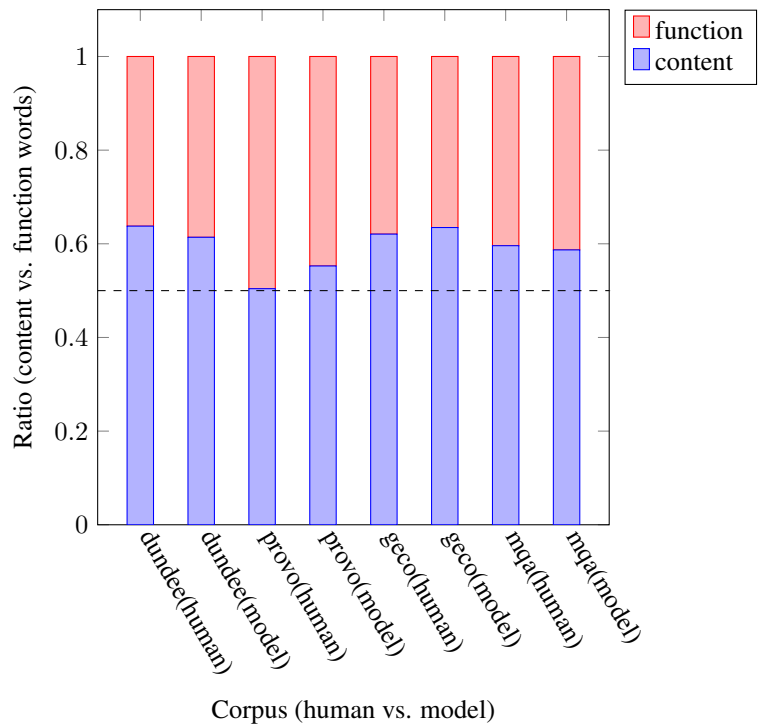

In our paper we showed that our model and humans are significantly correlated with respect to gaze durations over part of speech tag (POS) distributions.
We use this measure as POS tags have been shown to be good predictors of fixation probabilities~\cite{hahn2016modeling,nilsson2009learning}.
In Figure~\ref{fig:POS_distribution}, 
we provide an additional analysis on this matter.
We group together the fixation duration predictions over content words (adjective, adverb, noun, and verb) and the fixation duration predictions over function words (conjunction, pronoun, determiner, numbers, adposition, and particles), for both human gaze and our model predictions (normalized between 0 to 1). In the figure, we show that our model predicts, \textit{similarly to humans}, that content words are more informative than function words.

\end{document}

%% file: figures/1.tex



\begin{tikzpicture}

  \tikzstyle{neuron}=[draw, circle, thick, fill=white]
  \tikzstyle{layer}=[draw, rectangle, thick, fill=white, rounded corners, minimum height=1em, minimum width=14em]

  \node[neuron, very thick, xshift=28em, yshift=9em] (x) {};
  \node[left=.5em of x] (xh) {$\boldsymbol{x}$};


  \node (sal) [draw=none, minimum height=16em, minimum width=16em, xshift=8em, yshift=-4em, fill=olivegreen, fill opacity=0.2, very thick, rectangle, rounded corners] {};
  \node[below right= 1em of sal.north west, align=left] (tsm_label) {\textbf{Text Saliency}\\\textbf{Model}};

  \foreach \i in {1,...,7}
    \node[xshift=\i*2 em, yshift=4em] (x\i) {};
  \draw [-, thick, rounded corners=5pt] ($(x.south)+(0,-.9)$) -| (x4.north);

	\node (e) [layer, xshift=8em] {Embedding layer};
  \foreach \i in {1,...,7}
    \node[circle, minimum height=1.6em, xshift=\i*2 em] (eh\i) {};

  \draw[-stealth, thick] (x4.north) -- (eh4);

  \node (bil) [layer, xshift=8em, yshift=-3em] {BiLSTM layer};
  \foreach \i in {1,...,7}
    \node[circle, minimum height=1.6em, xshift=\i*2 em, yshift=-3em] (bilh\i) {};

  \foreach \x in {1,...,7}
    \draw[-stealth, thick] (eh\x) -- (bilh\x);

  \node (t) [layer, xshift=8em, yshift=-6em] {Transformer layer};
  \foreach \i in {1,...,7}
    \node[circle, minimum height=1.6em, xshift=\i*2 em, yshift=-6em] (th\i) {};

  \foreach \x in {1,...,7}
    \draw[-stealth, thick] (bilh\x) -- (th\x);

  \node (l) [layer, xshift=8em, yshift=-10em] {Output layer};
  \foreach \i in {1,...,7}
    \node[circle, minimum height=1.6em, xshift=\i*2 em, yshift=-10em] (lh\i) {};

  \foreach \x in {1,...,7}
    \foreach \y in {1,...,7}
      \draw[-stealth, thick] (th\x.south) -- (lh\y.north);

  \foreach \i in {1,...,7}
    \node[xshift=\i*2 em, yshift=-13.5em] (a\i) {};
  \node[left=.5em of a4] (a) {$\boldsymbol{u}$};

  \draw[-, very thick, red] (lh4) -- (a4);


  \node (task) [draw=none, minimum height=16em, minimum width=16em, xshift=28em, yshift=-4em, fill=blue, fill opacity=0.2, very thick, rectangle, rounded corners] {};
  \node[below right= 1em of task.north west, align=left] (tsm_label) {\textbf{Task Model}};

  \node (tl1) [layer, dashed, xshift=28em] {Task specific};
  \foreach \i in {1,...,7}
    \node[xshift=20em+\i*2 em, minimum height=1.6em] (tl1h\i) {};
  \draw [-, thick] (x.south) -- (tl1h4.north);

  \foreach \i in {1,...,7}
    \node[draw=none, minimum height=4.5em, xshift=20em+\i*2 em, yshift=0em] (m\i) {};

  \node (att) [draw, minimum height=3.5em, minimum width=18em, xshift=28em, yshift=-5em, fill=white, fill opacity=1.0, very thick, rectangle] {};
    
  \node (al) [layer, xshift=28em, yshift=-5em] {Attention layer};
  \foreach \i in {1,...,7}
    \node[minimum height=1.6em, xshift=20em+\i*2 em, yshift=-5em] (alh\i) {};

  \foreach \i in {1,...,7}
	  \draw[-stealth, thick] (tl1h\i) -- (alh\i);

  \draw [-stealth, very thick, red, rounded corners=5pt] (a4.north) |- ($(a4.south)+(3.5,0)$) |- ($(al.west)+(-.2,0)$);

  \node (tl2) [layer, dashed, xshift=28em, yshift=-10em] {Task specific};
  \foreach \i in {1,...,7}
    \node[xshift=20em+\i*2em, yshift=-10em, minimum height=1.6em] (tl2h\i) {};

  \foreach \i in {1,...,7}
    \draw[-latex, thick] (alh\i.south) -- (tl2h\i.north);

  
  \node[neuron, very thick, below=4em of tl2h4] (y) {};
  \node[left=.5em of y] (yh) {$\boldsymbol{y}$};
  \draw[-stealth, thick, shorten >=2pt] (tl2h4) -- (y);


  \node (att_zoom) [draw, minimum height=12em, minimum width=18em, xshift=48em, yshift=-5em, fill=white, fill opacity=0.2, very thick, rectangle, rounded corners] {};
  \draw[-, very thick, dashed] (att.north east) -- (att_zoom.north west);
  \draw[-, very thick, dashed] (att.south east) -- (att_zoom.south west);

  \foreach \i in {1,...,7}
    \node[xshift=41em+\i*2em, minimum height=1.1em, yshift=1.5em] (aih\i) {};

  \foreach \i in {1,...,7}
    \node[neuron, xshift=41em+\i*2em, yshift=-2em] (ash\i) {};

  \foreach \x in {1,...,7}
    \draw[-latex, thick] (aih\x) -- (ash\x);

  \node[draw, thick, rectangle, xshift=48em, yshift=-5em] (at) {$\boldsymbol{a_i}$};

  \foreach \i in {1,...,7}
    \node[neuron, xshift=41em+\i*2em, yshift=-8.5em] (ao\i) {};

  \foreach \i in {1,...,4}
    \draw[-latex, very thick, black!20, dashed] (ash\i) -- (ao\i);
  \foreach \i in {1,...,4}
    \draw[-latex, very thick, black!20, dashed] (at) -- (ao\i);
  \foreach \i in {6,...,7}
    \draw[-latex, very thick, black!20, dashed] (ash\i) -- (ao\i);
  \foreach \i in {6,...,7}
    \draw[-latex, very thick, black!20, dashed] (at) -- (ao\i);

  \node[xshift=38.6em, yshift=-5em] (fh) {};
    \draw[-latex, very thick, red] (fh) node[black, above, xshift=2em] {$\boldsymbol{u}$} -- (at);

  \foreach \x in {1,...,7}
    \draw[-latex, thick] (ash\x) -- (at.north);

  \foreach \i in {1,...,7}
    \node[xshift=41em+\i*2em, yshift=-11.4em] (aoh\i) {};

  \draw[-latex, very thick] (at) -- (ao5);
  \draw[-latex, very thick] (ash5) -- (ao5);
 
  \foreach \i in {1,...,7}
  \draw[-, thick] (ao\i) -- (aoh\i);

\end{tikzpicture}

%% file: neurips_2020.bbl
\begin{thebibliography}{87}
\providecommand{\natexlab}[1]{#1}
\providecommand{\url}[1]{\texttt{#1}}
\expandafter\ifx\csname urlstyle\endcsname\relax
  \providecommand{\doi}[1]{doi: #1}\else
  \providecommand{\doi}{doi: \begingroup \urlstyle{rm}\Url}\fi

\bibitem[Bahdanau et~al.(2015)Bahdanau, Cho, and Bengio]{bahdanau2014neural}
Dzmitry Bahdanau, Kyunghyun Cho, and Yoshua Bengio.
\newblock Neural machine translation by jointly learning to align and
  translate.
\newblock In \emph{Proc. International Conference on Learning Representations},
  2015.

\bibitem[Barrett et~al.(2018)Barrett, Bingel, Hollenstein, Rei, and
  S{\o}gaard]{barrett2018sequence}
Maria Barrett, Joachim Bingel, Nora Hollenstein, Marek Rei, and Anders
  S{\o}gaard.
\newblock Sequence classification with human attention.
\newblock In \emph{Proc. Conference on Computational Natural Language
  Learning}, pages 302--312, 2018.

\bibitem[Blohm et~al.(2018)Blohm, Jagfeld, Sood, Yu, and
  Vu]{blohm2018comparing}
Matthias Blohm, Glorianna Jagfeld, Ekta Sood, Xiang Yu, and Ngoc~Thang Vu.
\newblock Comparing attention-based convolutional and recurrent neural
  networks: Success and limitations in machine reading comprehension.
\newblock In \emph{Proc. Conference on Computational Natural Language
  Learning}, pages 108--118, 2018.
\newblock \doi{10.18653/v1/K18-1011}.

\bibitem[Borji and Itti(2012)]{borji2012state}
Ali Borji and Laurent Itti.
\newblock State-of-the-art in visual attention modeling.
\newblock \emph{IEEE transactions on pattern analysis and machine
  intelligence}, 35\penalty0 (1):\penalty0 185--207, 2012.

\bibitem[Britz et~al.(2017)Britz, Goldie, Luong, and Le]{britz2017massive}
Denny Britz, Anna Goldie, Minh-Thang Luong, and Quoc Le.
\newblock Massive exploration of neural machine translation architectures.
\newblock \emph{arXiv preprint arXiv:1703.03906}, 2017.

\bibitem[Bylinskii et~al.(2016)Bylinskii, Recasens, Borji, Oliva, Torralba, and
  Durand]{bylinskii2016should}
Zoya Bylinskii, Adri{\`a} Recasens, Ali Borji, Aude Oliva, Antonio Torralba,
  and Fr{\'e}do Durand.
\newblock Where should saliency models look next?
\newblock In \emph{Proc. European Conference on Computer Vision}, pages
  809--824. Springer, 2016.

\bibitem[Cao et~al.(2016)Cao, Li, Li, Wei, and Li]{cao2016attsum}
Ziqiang Cao, Wenjie Li, Sujian Li, Furu Wei, and Yanran Li.
\newblock {A}tt{S}um: Joint learning of focusing and summarization with neural
  attention.
\newblock In \emph{Proc. International Conference on Computational Linguistics:
  Technical Papers}, pages 547--556, December 2016.

\bibitem[Chen et~al.(2016)Chen, Bolton, and Manning]{chen2016thorough}
Danqi Chen, Jason Bolton, and Christopher~D. Manning.
\newblock A thorough examination of the {CNN}/daily mail reading comprehension
  task.
\newblock In \emph{Proc. Annual Meeting of the Association for Computational
  Linguistics}, pages 2358--2367, August 2016.
\newblock \doi{10.18653/v1/P16-1223}.

\bibitem[Cho et~al.(2019)Cho, Xie, and Campbell]{cho2019paraphrase}
Eunah Cho, He~Xie, and William~M Campbell.
\newblock Paraphrase generation for semi-supervised learning in nlu.
\newblock In \emph{Proc. Workshop on Methods for Optimizing and Evaluating
  Neural Language Generation}, pages 45--54, 2019.

\bibitem[Cho et~al.(2014)Cho, van Merri{\"e}nboer, Gulcehre, Bahdanau,
  Bougares, Schwenk, and Bengio]{cho2014learning}
Kyunghyun Cho, Bart van Merri{\"e}nboer, Caglar Gulcehre, Dzmitry Bahdanau,
  Fethi Bougares, Holger Schwenk, and Yoshua Bengio.
\newblock Learning phrase representations using {RNN} encoder{--}decoder for
  statistical machine translation.
\newblock In \emph{Proc. Conference on Empirical Methods in Natural Language
  Processing}, pages 1724--1734, 2014.
\newblock \doi{10.3115/v1/D14-1179}.

\bibitem[Cho et~al.(2015)Cho, Courville, and Bengio]{cho2015describing}
Kyunghyun Cho, Aaron Courville, and Yoshua Bengio.
\newblock Describing multimedia content using attention-based encoder-decoder
  networks.
\newblock \emph{IEEE Transactions on Multimedia}, 17\penalty0 (11):\penalty0
  1875--1886, 2015.

\bibitem[Clarke and Lapata(2008)]{clarke2008global}
James Clarke and Mirella Lapata.
\newblock Global inference for sentence compression: An integer linear
  programming approach.
\newblock \emph{Journal of Artificial Intelligence Research}, 31:\penalty0
  399--429, 2008.

\bibitem[Commodari and Guarnera(2005)]{commodari2005attention}
Elena Commodari and Maria Guarnera.
\newblock Attention and reading skills.
\newblock \emph{Perceptual and Motor Skills}, 100\penalty0 (2):\penalty0
  375--386, 2005.

\bibitem[Cop et~al.(2017)Cop, Dirix, Drieghe, and Duyck]{cop2017presenting}
Uschi Cop, Nicolas Dirix, Denis Drieghe, and Wouter Duyck.
\newblock Presenting geco: An eyetracking corpus of monolingual and bilingual
  sentence reading.
\newblock \emph{Behavior Research Methods}, 49\penalty0 (2):\penalty0 602--615,
  2017.

\bibitem[Davies et~al.(2016)Davies, Brown, Vigo, Harper, Horseman, Splendiani,
  Hill, and Jay]{davies2016exploring}
Alan Davies, Gavin Brown, Markel Vigo, Simon Harper, Laura Horseman, Bruno
  Splendiani, Elspeth Hill, and Caroline Jay.
\newblock Exploring the relationship between eye movements and
  electrocardiogram interpretation accuracy.
\newblock \emph{Scientific reports}, 6:\penalty0 38227, 2016.

\bibitem[Duch et~al.(2008)Duch, Oentaryo, and Pasquier]{duch2008cognitive}
W\l{}odzis\l{}aw Duch, Richard~J. Oentaryo, and Michel Pasquier.
\newblock Cognitive architectures: Where do we go from here?
\newblock In \emph{Proc. Conference on Artificial General Intelligence}, page
  122–136, NLD, 2008.
\newblock ISBN 9781586038335.

\bibitem[Engbert et~al.(2005)Engbert, Nuthmann, Richter, and
  Kliegl]{engbert2005swift}
Ralf Engbert, Antje Nuthmann, Eike~M Richter, and Reinhold Kliegl.
\newblock Swift: A dynamical model of saccade generation during reading.
\newblock \emph{Psychological Review}, 112\penalty0 (4):\penalty0 777, 2005.

\bibitem[Fader et~al.(2013)Fader, Zettlemoyer, and
  Etzioni]{fader2013paraphrase}
Anthony Fader, Luke Zettlemoyer, and Oren Etzioni.
\newblock Paraphrase-driven learning for open question answering.
\newblock In \emph{Proc. Meeting of the Association for Computational
  Linguistics (Volume 1: Long Papers)}, pages 1608--1618, 2013.

\bibitem[Fang et~al.(2009)Fang, Chai, and Ferreira]{fang2009between}
Rui Fang, Joyce~Y Chai, and Fernanda Ferreira.
\newblock Between linguistic attention and gaze fixations inmultimodal
  conversational interfaces.
\newblock In \emph{Proc. International Conference on Multimodal Interfaces},
  pages 143--150, 2009.

\bibitem[Filippova et~al.(2015)Filippova, Alfonseca, Colmenares, Kaiser, and
  Vinyals]{filippova2015sentence}
Katja Filippova, Enrique Alfonseca, Carlos~A Colmenares, {\L}ukasz Kaiser, and
  Oriol Vinyals.
\newblock Sentence compression by deletion with lstms.
\newblock In \emph{Proc. Conference on Empirical Methods in Natural Language
  Processing}, pages 360--368, 2015.

\bibitem[Frintrop et~al.(2010)Frintrop, Rome, and
  Christensen]{frintrop2010computational}
Simone Frintrop, Erich Rome, and Henrik~I Christensen.
\newblock Computational visual attention systems and their cognitive
  foundations: A survey.
\newblock \emph{ACM Transactions on Applied Perception (TAP)}, 7\penalty0
  (1):\penalty0 6, 2010.

\bibitem[Graves and Schmidhuber(2005)]{graves2005framewise}
Alex Graves and J{\"u}rgen Schmidhuber.
\newblock Framewise phoneme classification with bidirectional lstm and other
  neural network architectures.
\newblock \emph{Neural Networks}, 18\penalty0 (5-6):\penalty0 602--610, 2005.

\bibitem[Gupta et~al.(2018)Gupta, Agarwal, Singh, and Rai]{gupta2018deep}
Ankush Gupta, Arvind Agarwal, Prawaan Singh, and Piyush Rai.
\newblock A deep generative framework for paraphrase generation.
\newblock In \emph{Proc. Thirty-Second AAAI Conference on Artificial
  Intelligence}, 2018.

\bibitem[Gupta and Lehal(2010)]{gupta2010survey}
Vishal Gupta and Gurpreet~Singh Lehal.
\newblock A survey of text summarization extractive techniques.
\newblock \emph{Journal of Emerging Technologies in Web Intelligence},
  2\penalty0 (3):\penalty0 258--268, 2010.

\bibitem[Hahn and Keller(2016)]{hahn2016modeling}
Michael Hahn and Frank Keller.
\newblock Modeling human reading with neural attention.
\newblock In \emph{Proc. Conference on Empirical Methods in Natural Language
  Processing}, pages 85--95, 2016.
\newblock \doi{10.18653/v1/D16-1009}.

\bibitem[Hahn and Keller(2018)]{DBLP:journals/corr/abs-1808-00054}
Michael Hahn and Frank Keller.
\newblock Modeling task effects in human reading with neural attention.
\newblock \emph{CoRR}, abs/1808.00054, 2018.
\newblock URL \url{http://arxiv.org/abs/1808.00054}.

\bibitem[Hasan et~al.(2016)Hasan, Liu, Liu, Qadir, Lee, Datla, Prakash, and
  Farri]{hasan2016neural}
Sadid~A Hasan, Bo~Liu, Joey Liu, Ashequl Qadir, Kathy Lee, Vivek Datla, Aaditya
  Prakash, and Oladimeji Farri.
\newblock Neural clinical paraphrase generation with attention.
\newblock In \emph{Proc. Clinical Natural Language Processing Workshop}, pages
  42--53, 2016.

\bibitem[Hermann et~al.(2015)Hermann, Kocisky, Grefenstette, Espeholt, Kay,
  Suleyman, and Blunsom]{hermann2015teaching}
Karl~Moritz Hermann, Tomas Kocisky, Edward Grefenstette, Lasse Espeholt, Will
  Kay, Mustafa Suleyman, and Phil Blunsom.
\newblock Teaching machines to read and comprehend.
\newblock In \emph{Proc. Advances in Neural Information Processing Systems},
  pages 1693--1701, 2015.

\bibitem[Hollenstein et~al.(2019)Hollenstein, Barrett, Troendle, Bigiolli,
  Langer, and Zhang]{hollenstein2019advancing}
Nora Hollenstein, Maria Barrett, Marius Troendle, Francesco Bigiolli, Nicolas
  Langer, and Ce~Zhang.
\newblock Advancing nlp with cognitive language processing signals.
\newblock \emph{arXiv preprint arXiv:1904.02682}, 2019.

\bibitem[Iida et~al.(2011)Iida, Yasuhara, and Tokunaga]{iida2011multi}
Ryu Iida, Masaaki Yasuhara, and Takenobu Tokunaga.
\newblock Multi-modal reference resolution in situated dialogue by integrating
  linguistic and extra-linguistic clues.
\newblock In \emph{Proc. International Joint Conference on Natural Language
  Processing}, pages 84--92, 2011.

\bibitem[Jing(2000)]{jing2000sentence}
Hongyan Jing.
\newblock Sentence reduction for automatic text summarization.
\newblock In \emph{Proc. Applied Natural Language Processing Conference}, pages
  310--315, 2000.

\bibitem[Karessli et~al.(2017)Karessli, Akata, Schiele, and
  Bulling]{karessli2017gaze}
Nour Karessli, Zeynep Akata, Bernt Schiele, and Andreas Bulling.
\newblock Gaze embeddings for zero-shot image classification.
\newblock In \emph{Proc. IEEE Conference on Computer Vision and Pattern
  Recognition}, pages 4525--4534, 2017.

\bibitem[Kaushik and Lipton(2018)]{kaushik2018much}
Divyansh Kaushik and Zachary~C Lipton.
\newblock How much reading does reading comprehension require? a critical
  investigation of popular benchmarks.
\newblock In \emph{Proc. Conference on Empirical Methods in Natural Language
  Processing}, pages 5010--5015, 2018.

\bibitem[Kennedy and Pynte(2005)]{kennedy2005parafoveal}
Alan Kennedy and Jo{\"e}l Pynte.
\newblock Parafoveal-on-foveal effects in normal reading.
\newblock \emph{Vision research}, 45\penalty0 (2):\penalty0 153--168, 2005.

\bibitem[Kingma and Ba(2015)]{kingma2014adam}
Diederik~P. Kingma and Jimmy Ba.
\newblock Adam: {A} method for stochastic optimization.
\newblock In \emph{Proc. International Conference on Learning Representations},
  2015.

\bibitem[Klerke and Plank(2019)]{klerke2019glance}
Sigrid Klerke and Barbara Plank.
\newblock At a glance: The impact of gaze aggregation views on syntactic
  tagging.
\newblock In \emph{Proc. Beyond Vision and Language: Integrating Real-World
  Knowledge}, pages 51--61, 2019.

\bibitem[Klerke et~al.(2016)Klerke, Goldberg, and
  S{\o}gaard]{klerke2016improving}
Sigrid Klerke, Yoav Goldberg, and Anders S{\o}gaard.
\newblock Improving sentence compression by learning to predict gaze.
\newblock In \emph{Proc. Conference of the North American Chapter of the
  Association for Computational Linguistics}, pages 1528--1533, 2016.
\newblock \doi{10.18653/v1/N16-1179}.

\bibitem[Knight and Marcu(2002)]{knight2002summarization}
Kevin Knight and Daniel Marcu.
\newblock Summarization beyond sentence extraction: A probabilistic approach to
  sentence compression.
\newblock \emph{Artificial Intelligence}, 139\penalty0 (1):\penalty0 91--107,
  2002.

\bibitem[Kotseruba and Tsotsos(2018)]{kotseruba201840}
Iuliia Kotseruba and John~K Tsotsos.
\newblock 40 years of cognitive architectures: core cognitive abilities and
  practical applications.
\newblock \emph{Artificial Intelligence Review}, pages 1--78, 2018.

\bibitem[K{\"u}mmerer et~al.(2015)K{\"u}mmerer, Theis, and
  Bethge]{kummerer2014deep}
Matthias K{\"u}mmerer, Lucas Theis, and Matthias Bethge.
\newblock Deep gaze i: Boosting saliency prediction with feature maps trained
  on imagenet.
\newblock In \emph{In International Conference on Learning Representations},
  pages 1--12, 2015.

\bibitem[Li et~al.(2018)Li, Jiang, Shang, and Li]{li2017paraphrase}
Zichao Li, Xin Jiang, Lifeng Shang, and Hang Li.
\newblock Paraphrase generation with deep reinforcement learning.
\newblock In \emph{Proc. Conference on Empirical Methods in Natural Language
  Processing}, pages 3865--3878, October-November 2018.
\newblock \doi{10.18653/v1/D18-1421}.

\bibitem[Lin(1991)]{lin1991divergence}
Jianhua Lin.
\newblock Divergence measures based on the shannon entropy.
\newblock \emph{IEEE Transactions on Information theory}, 37\penalty0
  (1):\penalty0 145--151, 1991.

\bibitem[Luke and Christianson(2018)]{luke2018provo}
Steven~G Luke and Kiel Christianson.
\newblock The provo corpus: A large eye-tracking corpus with predictability
  norms.
\newblock \emph{Behavior Research Methods}, 50\penalty0 (2):\penalty0 826--833,
  2018.

\bibitem[Luong et~al.(2015)Luong, Pham, and
  Manning]{DBLP:journals/corr/LuongPM15}
Minh-Thang Luong, Hieu Pham, and Christopher~D Manning.
\newblock Effective approaches to attention-based neural machine translation.
\newblock In \emph{Proc. Conference on Empirical Methods in Natural Language
  Processing}, pages 1412--1421, 2015.

\bibitem[Ma and Peters(2020)]{ma2020neural}
Wei~Ji Ma and Benjamin Peters.
\newblock A neural network walks into a lab: towards using deep nets as models
  for human behavior.
\newblock \emph{arXiv preprint arXiv:2005.02181}, 2020.

\bibitem[Matthies and S{\o}gaard(2013)]{matthies2013blinkers}
Franz Matthies and Anders S{\o}gaard.
\newblock With blinkers on: Robust prediction of eye movements across readers.
\newblock In \emph{Proc. Conference on Empirical Methods in Natural Language
  Processing}, pages 803--807, 2013.

\bibitem[McDonald(2006)]{mcdonald2006discriminative}
Ryan McDonald.
\newblock Discriminative sentence compression with soft syntactic evidence.
\newblock In \emph{Proc. Conference of the European Chapter of the Association
  for Computational Linguistics}, 2006.

\bibitem[Mnih et~al.(2014)Mnih, Heess, Graves, et~al.]{mnih2014recurrent}
Volodymyr Mnih, Nicolas Heess, Alex Graves, et~al.
\newblock Recurrent models of visual attention.
\newblock In \emph{Proc. Advances in Neural Information Processing Systems},
  pages 2204--2212, 2014.

\bibitem[Mozaffari et~al.(2018)Mozaffari, Klein, Viiri, Ahmed, Kuhn, and
  Dengel]{mozaffari2018evaluating}
Saleh Mozaffari, Pascal Klein, Jouni Viiri, Sheraz Ahmed, Jochen Kuhn, and
  Andreas Dengel.
\newblock Evaluating similarity measures for gaze patterns in the context of
  representational competence in physics education.
\newblock In \emph{Proc. ACM Symposium on Eye Tracking Research \&
  Applications}, pages 1--5, 2018.

\bibitem[Nilsson and Nivre(2009)]{nilsson2009learning}
Mattias Nilsson and Joakim Nivre.
\newblock Learning where to look: Modeling eye movements in reading.
\newblock In \emph{Proc. Conference on Computational Natural Language
  Learning}, pages 93--101, 2009.

\bibitem[Nilsson and Nivre(2010)]{nilsson2010towards}
Mattias Nilsson and Joakim Nivre.
\newblock Towards a data-driven model of eye movement control in reading.
\newblock In \emph{Proc. Workshop on Cognitive Modeling and Computational
  Linguistics}, pages 63--71. Association for Computational Linguistics, 2010.

\bibitem[Oertel and Salvi(2013)]{oertel2013gaze}
Catharine Oertel and Giampiero Salvi.
\newblock A gaze-based method for relating group involvement to individual
  engagement in multimodal multiparty dialogue.
\newblock In \emph{Proc. International Conference on Multimodal Interaction},
  pages 99--106, 2013.

\bibitem[Papineni et~al.(2002)Papineni, Roukos, Ward, and
  Zhu]{papineni2002bleu}
Kishore Papineni, Salim Roukos, Todd Ward, and Wei-Jing Zhu.
\newblock Bleu: a method for automatic evaluation of machine translation.
\newblock In \emph{Proc. Annual Meeting of Association for Computational
  Linguistics}, pages 311--318, 2002.

\bibitem[Patro et~al.(2018)Patro, Kurmi, Kumar, and
  Namboodiri]{patro2018learning}
Badri~Narayana Patro, Vinod~Kumar Kurmi, Sandeep Kumar, and Vinay Namboodiri.
\newblock Learning semantic sentence embeddings using sequential pair-wise
  discriminator.
\newblock In \emph{Proc. International Conference on Computational
  Linguistics}, pages 2715--2729, 2018.

\bibitem[Pennington et~al.(2014)Pennington, Socher, and
  Manning]{pennington2014glove}
Jeffrey Pennington, Richard Socher, and Christopher~D Manning.
\newblock Glove: Global vectors for word representation.
\newblock In \emph{Proc. Conference on Empirical Methods in Natural Language
  Processing}, pages 1532--1543, 2014.

\bibitem[Post(2018)]{post2018call}
Matt Post.
\newblock A call for clarity in reporting bleu scores.
\newblock In \emph{Proc. Conference on Machine Translation}, pages 186--191,
  2018.

\bibitem[Prakash et~al.(2016)Prakash, Hasan, Lee, Datla, Qadir, Liu, and
  Farri]{prakash2016neural}
Aaditya Prakash, Sadid~A. Hasan, Kathy Lee, Vivek Datla, Ashequl Qadir, Joey
  Liu, and Oladimeji Farri.
\newblock Neural paraphrase generation with stacked residual {LSTM} networks.
\newblock In \emph{Proc. International Conference on Computational
  Linguistics}, pages 2923--2934, December 2016.

\bibitem[Qiao et~al.(2018)Qiao, Dong, and Xu]{qiao2018exploring}
Tingting Qiao, Jianfeng Dong, and Duanqing Xu.
\newblock Exploring human-like attention supervision in visual question
  answering.
\newblock In \emph{Proc. Thirty-Second AAAI Conference on Artificial
  Intelligence}, 2018.

\bibitem[Rayner(1978)]{rayner1978eye}
Keith Rayner.
\newblock Eye movements in reading and information processing.
\newblock \emph{Psychological bulletin}, 85\penalty0 (3):\penalty0 618, 1978.

\bibitem[Reichle et~al.(1998)Reichle, Pollatsek, Fisher, and
  Rayner]{reichle1998toward}
Erik~D Reichle, Alexander Pollatsek, Donald~L Fisher, and Keith Rayner.
\newblock Toward a model of eye movement control in reading.
\newblock \emph{Psychological review}, 105\penalty0 (1):\penalty0 125, 1998.

\bibitem[Reichle et~al.(2009)Reichle, Warren, and McConnell]{reichle2009using}
Erik~D Reichle, Tessa Warren, and Kerry McConnell.
\newblock Using ez reader to model the effects of higher level language
  processing on eye movements during reading.
\newblock \emph{Psychonomic bulletin \& review}, 16\penalty0 (1):\penalty0
  1--21, 2009.

\bibitem[Reichle et~al.(2013)Reichle, Liversedge, Drieghe, Blythe, Joseph,
  White, and Rayner]{reichle2013using}
Erik~D Reichle, Simon~P Liversedge, Denis Drieghe, Hazel~I Blythe, Holly~SSL
  Joseph, Sarah~J White, and Keith Rayner.
\newblock Using ez reader to examine the concurrent development of eye-movement
  control and reading skill.
\newblock \emph{Developmental Review}, 33\penalty0 (2):\penalty0 110--149,
  2013.

\bibitem[Rockt{\"{a}}schel et~al.(2016)Rockt{\"{a}}schel, Grefenstette,
  Hermann, Kocisk{\'{y}}, and Blunsom]{rocktaschel2015reasoning}
Tim Rockt{\"{a}}schel, Edward Grefenstette, Karl~Moritz Hermann, Tom{\'{a}}s
  Kocisk{\'{y}}, and Phil Blunsom.
\newblock Reasoning about entailment with neural attention.
\newblock In \emph{Proc. International Conference on Learning Representations},
  2016.

\bibitem[Rohanian et~al.(2017)Rohanian, Taslimipoor, Yaneva, and
  Ha]{rohanian2017using}
Omid Rohanian, Shiva Taslimipoor, Victoria Yaneva, and Le~An Ha.
\newblock Using gaze data to predict multiword expressions.
\newblock In \emph{Proc. International Conference Recent Advances in Natural
  Language Processing}, pages 601--609, Varna, Bulgaria, September 2017.
\newblock \doi{10.26615/978-954-452-049-6_078}.

\bibitem[Rush et~al.(2015)Rush, Chopra, and Weston]{rush2015neural}
Alexander~M. Rush, Sumit Chopra, and Jason Weston.
\newblock A neural attention model for abstractive sentence summarization.
\newblock In \emph{Proc. Conference on Empirical Methods in Natural Language
  Processing}, pages 379--389, September 2015.
\newblock \doi{10.18653/v1/D15-1044}.

\bibitem[Samardzhiev et~al.(2018)Samardzhiev, Gargett, and
  Bollegala]{samardzhiev2018learning}
Krasen Samardzhiev, Andrew Gargett, and Danushka Bollegala.
\newblock Learning neural word salience scores.
\newblock In \emph{Proc. Joint Conference on Lexical and Computational
  Semantics}, pages 33--42, 2018.
\newblock \doi{10.18653/v1/S18-2004}.

\bibitem[Schuster and Paliwal(1997)]{schuster1997bidirectional}
Mike Schuster and Kuldip~K Paliwal.
\newblock Bidirectional recurrent neural networks.
\newblock \emph{IEEE transactions on Signal Processing}, 45\penalty0
  (11):\penalty0 2673--2681, 1997.

\bibitem[Seo et~al.(2018)Seo, Min, Farhadi, and Hajishirzi]{seo2017neural}
Minjoon Seo, Sewon Min, Ali Farhadi, and Hannaneh Hajishirzi.
\newblock Neural speed reading via skim-{RNN}.
\newblock In \emph{Proc. International Conference on Learning Representations},
  2018.

\bibitem[Shcherbatyi et~al.(2015)Shcherbatyi, Bulling, and
  Fritz]{shcherbatyi2015gazedpm}
Iaroslav Shcherbatyi, Andreas Bulling, and Mario Fritz.
\newblock Gazedpm: Early integration of gaze information in deformable part
  models.
\newblock \emph{arXiv preprint arXiv:1505.05753}, 2015.

\bibitem[Sood et~al.(2020)Sood, Tannert, Frassinelli, Bulling, and
  Vu]{sood20_conll}
Ekta Sood, Simon Tannert, Diego Frassinelli, Andreas Bulling, and Ngoc~Thang
  Vu.
\newblock Interpreting attention models with human visual attention in machine
  reading comprehension.
\newblock In \emph{Proc. ACL SIGNLL Conference on Computational Natural
  Language Learning (CoNLL)}, 2020.

\bibitem[Sugano and Bulling(2016)]{sugano2016seeing}
Yusuke Sugano and Andreas Bulling.
\newblock Seeing with humans: Gaze-assisted neural image captioning.
\newblock \emph{arXiv preprint arXiv:1608.05203}, 2016.

\bibitem[Tapaswi et~al.(2016)Tapaswi, Zhu, Stiefelhagen, Torralba, Urtasun, and
  Fidler]{MovieQA}
Makarand Tapaswi, Yukun Zhu, Rainer Stiefelhagen, Antonio Torralba, Raquel
  Urtasun, and Sanja Fidler.
\newblock {MovieQA: Understanding Stories in Movies through
  Question-Answering}.
\newblock In \emph{Proc. IEEE Conference on Computer Vision and Pattern
  Recognition}, 2016.

\bibitem[Tas and Kiyani(2007)]{tas2007survey}
Oguzhan Tas and Farzad Kiyani.
\newblock A survey automatic text summarization.
\newblock \emph{PressAcademia Procedia}, 5\penalty0 (1):\penalty0 205--213,
  2007.

\bibitem[Toutanova et~al.(2003)Toutanova, Klein, Manning, and
  Singer]{10.3115/1073445.1073478}
Kristina Toutanova, Dan Klein, Christopher~D. Manning, and Yoram Singer.
\newblock Feature-rich part-of-speech tagging with a cyclic dependency network.
\newblock In \emph{Proc. Conference of the North American Chapter of the
  Association for Computational Linguistics}, page 173–180, 2003.
\newblock \doi{10.3115/1073445.1073478}.

\bibitem[Vaswani et~al.(2017)Vaswani, Shazeer, Parmar, Uszkoreit, Jones, Gomez,
  Kaiser, and Polosukhin]{vaswani2017attention}
Ashish Vaswani, Noam Shazeer, Niki Parmar, Jakob Uszkoreit, Llion Jones,
  Aidan~N Gomez, {\L}ukasz Kaiser, and Illia Polosukhin.
\newblock Attention is all you need.
\newblock In \emph{Proc. Advances in Neural Information Processing Systems},
  pages 5998--6008, 2017.

\bibitem[Wang et~al.(2017)Wang, Zhang, and Zong]{wang2016learning}
Shaonan Wang, Jiajun Zhang, and Chengqing Zong.
\newblock Learning sentence representation with guidance of human attention.
\newblock In \emph{Proc. International Joint Conference on Artificial
  Intelligence}, pages 4137--4143, 2017.

\bibitem[Wang et~al.(2019)Wang, Ma, Liu, and Tang]{wang2019r}
Zhiwei Wang, Yao Ma, Zitao Liu, and Jiliang Tang.
\newblock R-transformer: Recurrent neural network enhanced transformer.
\newblock \emph{arXiv:1907.05572}, 2019.

\bibitem[Wolfe and Horowitz(2017)]{wolfe2017five}
Jeremy~M Wolfe and Todd~S Horowitz.
\newblock Five factors that guide attention in visual search.
\newblock \emph{Nature Human Behaviour}, 1\penalty0 (3):\penalty0 1--8, 2017.

\bibitem[Xia et~al.(2019)Xia, Kochmar, and Briscoe]{xia2019automatic}
Menglin Xia, Ekaterina Kochmar, and Ted Briscoe.
\newblock Automatic learner summary assessment for reading comprehension.
\newblock In \emph{Proc. Conference of the North American Chapter of the
  Association for Computational Linguistics}, pages 2532--2542, 2019.
\newblock \doi{10.18653/v1/N19-1261}.

\bibitem[Xu et~al.(2015{\natexlab{a}})Xu, Mukherjee, Li, Warner, Rehg, and
  Singh]{xu2015gaze}
Jia Xu, Lopamudra Mukherjee, Yin Li, Jamieson Warner, James~M Rehg, and Vikas
  Singh.
\newblock Gaze-enabled egocentric video summarization via constrained
  submodular maximization.
\newblock In \emph{Proc. IEEE Conference on Computer Vision and Pattern
  Recognition}, pages 2235--2244, 2015{\natexlab{a}}.

\bibitem[Xu et~al.(2015{\natexlab{b}})Xu, Ba, Kiros, Cho, Courville,
  Salakhudinov, Zemel, and Bengio]{xu2015show}
Kelvin Xu, Jimmy Ba, Ryan Kiros, Kyunghyun Cho, Aaron Courville, Ruslan
  Salakhudinov, Rich Zemel, and Yoshua Bengio.
\newblock Show, attend and tell: Neural image caption generation with visual
  attention.
\newblock In \emph{Proc. International Conference on Machine Learning}, pages
  2048--2057, 2015{\natexlab{b}}.

\bibitem[Yaneva et~al.(2018)Yaneva, Ha, Evans, and
  Mitkov]{yaneva2018classifying}
Victoria Yaneva, Le~An Ha, Richard Evans, and Ruslan Mitkov.
\newblock Classifying referential and non-referential it using gaze.
\newblock In \emph{Proc. Conference on Empirical Methods in Natural Language
  Processing}, pages 4896--4901, Brussels, Belgium, October-November 2018.
\newblock \doi{10.18653/v1/D18-1528}.

\bibitem[Yu et~al.(2017)Yu, Choi, Kim, Yoo, Lee, and Kim]{yu2017supervising}
Youngjae Yu, Jongwook Choi, Yeonhwa Kim, Kyung Yoo, Sang-Hun Lee, and Gunhee
  Kim.
\newblock Supervising neural attention models for video captioning by human
  gaze data.
\newblock In \emph{Proc. IEEE Conference on Computer Vision and Pattern
  Recognition}, pages 490--498, 2017.

\bibitem[Yun et~al.(2013)Yun, Peng, Samaras, Zelinsky, and
  Berg]{yun2013studying}
Kiwon Yun, Yifan Peng, Dimitris Samaras, Gregory~J Zelinsky, and Tamara~L Berg.
\newblock Studying relationships between human gaze, description, and computer
  vision.
\newblock In \emph{Proc. IEEE Conference on Computer Vision and Pattern
  Recognition}, pages 739--746, 2013.

\bibitem[Zhang et~al.(2018)Zhang, Liu, Liu, Gao, Duh, and
  Van~Durme]{zhang2018record}
Sheng Zhang, Xiaodong Liu, Jingjing Liu, Jianfeng Gao, Kevin Duh, and Benjamin
  Van~Durme.
\newblock Record: Bridging the gap between human and machine commonsense
  reading comprehension.
\newblock \emph{arXiv preprint arXiv:1810.12885}, 2018.

\bibitem[Zhang and Zhang(2019)]{zhang2019using}
Yingyi Zhang and Chengzhi Zhang.
\newblock Using human attention to extract keyphrase from microblog post.
\newblock In \emph{Proc. Annual Meeting of the Association for Computational
  Linguistics}, pages 5867--5872, 2019.

\bibitem[Zhao et~al.(2018)Zhao, Luo, and Aizawa]{zhao2018language}
Yang Zhao, Zhiyuan Luo, and Akiko Aizawa.
\newblock A language model based evaluator for sentence compression.
\newblock In \emph{Proc. Annual Meeting of the Association for Computational
  Linguistics}, pages 170--175, 2018.

\end{thebibliography}
